\def\swordtitleruleplacement{1}
\def\swordtitlebrandvariant{2}
\def\swordtitlefontvariant{1}
\documentclass[oneside, a4paper, onecolumn, 10pt]{article}

\usepackage{booktabs}
\usepackage{array}
\usepackage{amssymb}
\usepackage{tabularx}

\title{DEPICT: Scoring Text-to-Image Alignment by Answer Agreement}

\author{%
{\normalsize
\textbf{Vasco Ramos}\textsuperscript{1,\,2,\,}\footnote[1]{This work was done during Vasco’s internship at Sword.},\enspace \textbf{Sandra Godinho Silva}\textsuperscript{1},\enspace 
\textbf{Joao Magalhaes}\textsuperscript{2},\enspace 
\textbf{Ricardo Rei}\textsuperscript{1},\enspace 
\textbf{Pedro Henrique Martins}\textsuperscript{1}}\\[1.1mm]
{\small \textsuperscript{1}Sword Health}\\[0.45mm]
{\small \textsuperscript{2}NOVALINCS, NOVA University of Lisbon}\\[0.45mm]
{\small \href{mailto:ai.research@sword.com}{\textcolor{black}{{\footnotesize\faEnvelope}\hspace{0.45em}ai.research@sword.com}}}
}
\date{}
\usepackage[dvipsnames]{xcolor}

\definecolor{linkblue}{RGB}{0,82,155}
\definecolor{citegreen}{RGB}{0,110,70}
\definecolor{urlviolet}{RGB}{100,40,130}

\definecolor{swordcoral}{RGB}{194,64,47}
\definecolor{swordblue}{RGB}{58,84,176}
\definecolor{swordviolet}{RGB}{105,74,158}

\definecolor{lstbg}{RGB}{250,246,239}
\definecolor{lstframe}{RGB}{223,220,213}
\definecolor{lstcmt}{RGB}{0,110,70}
\definecolor{lstrole}{RGB}{0,82,155}
\definecolor{lststr}{RGB}{60,60,60}
\definecolor{lstkw}{RGB}{140,30,120}
\definecolor{hlcol}{RGB}{250,246,239}

\definecolor{tokfg}{HTML}{374151}
\definecolor{tokbg}{HTML}{F3F4F6}

\definecolor{bordo}{RGB}{128,0,32}

\definecolor{panelPrompt}{HTML}{FCF3EA}
\definecolor{panelReason}{HTML}{F4F4F4}
\definecolor{panelAnswer}{HTML}{E8E8E4}
\definecolor{panelBorder}{HTML}{B9B9B9}
\definecolor{panelInk}{HTML}{27272A}
\definecolor{panelInkStrong}{HTML}{141414}
\definecolor{panelAccent}{HTML}{C43C45}
\definecolor{panelAccentBG}{HTML}{F3DEDE}

\usepackage{fullpage}
\usepackage[left=2.5cm,top=1.7cm,bottom=2.2cm,right=2.5cm]{geometry}
\usepackage[utf8]{inputenc}
\usepackage[T2A,T1]{fontenc}
\usepackage{iftex}
\ifPDFTeX
  \usepackage{CJKutf8}
  \DeclareUnicodeCharacter{211D}{\ensuremath{\mathbb{R}}}
  \DeclareUnicodeCharacter{2202}{\ensuremath{\partial}}
  \DeclareUnicodeCharacter{2212}{\ensuremath{-}}
\fi

\usepackage{amsmath}
\usepackage{amssymb}
\usepackage{bbm}
\usepackage{bm}

\usepackage{booktabs}
\usepackage{multirow}
\usepackage{colortbl}

\usepackage{graphicx}
\usepackage{wrapfig}
\usepackage{needspace}
\usepackage{float}
\usepackage[most]{tcolorbox}

\usepackage{pifont}
\usepackage{fontawesome5}
\usepackage{soul}
\usepackage{eurosym}
\usepackage{xspace}
\usepackage{seqsplit}
\usepackage{comment}
\usepackage{enumitem}

\usepackage{tgpagella}
\usepackage{roboto}
\renewcommand{\sfdefault}{cmss}%
\usepackage{nopageno}

\usepackage[authoryear,round]{natbib}

\usepackage[
  colorlinks=true,
  linkcolor=swordblue,
  citecolor=swordblue,
  urlcolor=swordblue,
  unicode
]{hyperref}
\usepackage{xurl}
\usepackage{footnotehyper}
\makesavenoteenv{figure}

\usepackage{etoc}
\usepackage[nameinlink,capitalise,noabbrev]{cleveref}

\usepackage{siunitx}

\usepackage{tikz}
\usetikzlibrary{positioning, arrows.meta, calc, decorations.pathreplacing, fit, backgrounds, patterns}
\usepackage{pgfplots}
\usepgfplotslibrary{groupplots}
\pgfplotsset{compat=1.16}
\usepackage{fancyvrb}
\usepackage{fvextra}
\usepackage{algorithm}
\usepackage{algpseudocode}
\usepackage{listings}

\usepackage{iftex}
\ifPDFTeX
  \newcommand{\bodyfont}{\fontfamily{XCharter-TLF}\selectfont}
  \newcommand{\labelfont}{\fontfamily{phv}\selectfont}
  \newcommand{\paperromanfont}{\normalfont}
\else
  \usepackage{fontspec}
  \newfontfamily\bodyfont{Spectral}[
    Path=fonts/, Extension=.ttf,
    UprightFont=*-Regular, ItalicFont=*-Italic,
    BoldFont=*-Bold, BoldItalicFont=*-BoldItalic]
  \newfontfamily\labelfont{LibreFranklin}[
    Path=fonts/, Extension=.ttf,
    UprightFont=*-SemiBold, BoldFont=*-Bold]
  \newfontfamily\paperromanfont{lmroman10-regular.otf}[Ligatures=TeX]
\fi

\ifPDFTeX
  
\else
  \newfontfamily\intlfont{NotoSansMonoCJKsc-Regular.otf}[Path=fonts/, Scale=MatchLowercase]
  
  \IfFontExistsTF{Noto Sans Devanagari}
    {\newfontfamily\devfont{Noto Sans Devanagari}[Scale=MatchLowercase]}{\let\devfont\intlfont}
  \IfFontExistsTF{Noto Sans Georgian}
    {\newfontfamily\geofont{Noto Sans Georgian}[Scale=MatchLowercase]}{\let\geofont\intlfont}

\fi
\ifPDFTeX \fi

\lstdefinestyle{trace}{
  basicstyle=\ttfamily\scriptsize,
  backgroundcolor=\color{lstbg},
  frame=single, rulecolor=\color{lstframe}, framerule=0.6pt,
  breaklines=true, breakindent=0pt, postbreak=\mbox{\textcolor{lstframe}{$\hookrightarrow$}\space},
  columns=fullflexible, keepspaces=true,
  commentstyle=\color{lstcmt}\itshape,
  keywordstyle=\color{lstrole}\bfseries,
  emphstyle=\color{lstrole}\bfseries,
  morekeywords={messages,user,assistant,system,thinking,text,signature,primer,decoder,reconciler},
  morecomment=[l]{\#},
  aboveskip=2pt, belowskip=2pt,
  xleftmargin=5pt, xrightmargin=3pt, framexleftmargin=5pt,
}

\lstdefinestyle{pyclean}{
  language=Python,
  basicstyle=\ttfamily\scriptsize,
  backgroundcolor=\color{lstbg},
  frame=single, rulecolor=\color{lstframe}, framerule=0.6pt,
  breaklines=true, breakindent=0pt,
  postbreak=\mbox{\textcolor{lstframe}{$\hookrightarrow$}\space},
  columns=fullflexible, keepspaces=true,
  showstringspaces=false,
  commentstyle=\color{lstcmt}\itshape,
  keywordstyle=\color{lstkw}\bfseries,
  stringstyle=\color{lstrole},
  emphstyle=\color{lstrole}\bfseries,
  emph={client,messages,create,model,max_tokens,thinking,content,role,signature},
  aboveskip=6pt, belowskip=6pt,
  xleftmargin=5pt, xrightmargin=3pt, framexleftmargin=5pt,
}

\lstdefinestyle{extracttemplate}{
  basicstyle=\ttfamily\footnotesize,
  frame=none,
  backgroundcolor={},
  breaklines=true,
  breakindent=12pt,
  columns=fullflexible,
  keepspaces=true,
  showstringspaces=false,
  alsoletter={@},
  morekeywords={@thought},
  keywordstyle=\color{panelAccent}\bfseries,
  commentstyle=\color{panelAccent}\itshape,
  morecomment=[l]{//},
  xleftmargin=0pt,
  xrightmargin=0pt,
  aboveskip=0pt,
  belowskip=0pt,
}

\lstdefinestyle{extractcall}{
  language=Python,
  basicstyle=\ttfamily\scriptsize,
  frame=none,
  backgroundcolor={},
  breaklines=true,
  breakindent=12pt,
  columns=fullflexible,
  keepspaces=true,
  showstringspaces=false,
  commentstyle=\color{lstcmt}\itshape,
  keywordstyle=\color{lstkw}\bfseries,
  stringstyle=\color{lstrole},
  emphstyle=\color{panelAccent}\bfseries,
  emph={client,messages,create,generate_content,responses,models,Content,Part},
  xleftmargin=0pt,
  xrightmargin=0pt,
  aboveskip=0pt,
  belowskip=0pt,
}

\lstdefinestyle{judgeprompt}{
  basicstyle=\ttfamily\scriptsize,
  frame=none,
  backgroundcolor={},
  breaklines=true,
  breakindent=12pt,
  columns=fullflexible,
  keepspaces=true,
  showstringspaces=false,
  alsoletter={_-},
  emph={META,SUB,REAL,NOT-ARTIFACT,RESOLUTION,Reasoning trace},
  emphstyle=\color{panelAccent}\bfseries,
  xleftmargin=0pt,
  xrightmargin=0pt,
  aboveskip=0pt,
  belowskip=0pt,
}
\usepackage{xcolor}
\definecolor{accent}{HTML}{C43C45}

\usepackage[font=small,labelfont=bf,skip=4pt,format=plain]{caption}
\usepackage{fancyhdr}
\crefname{appendix}{Appendix}{Appendices}
\Crefname{appendix}{Appendix}{Appendices}
\crefname{app}{Appendix}{Appendices}
\Crefname{app}{Appendix}{Appendices}

\setitemize{itemsep=4pt,topsep=-3pt,parsep=0pt,partopsep=0pt,leftmargin=18pt}

\makeatletter
\renewcommand{\paragraph}{%
  \@startsection{paragraph}{4}%
  {\z@}{0.5ex \@plus 1ex \@minus .2ex}{-1em}%
  {\normalfont\normalsize\bfseries}%
}
\makeatother

\makeatletter
\let\origsection\section
\renewcommand\section{\@ifstar{\starsection}{\nostarsection}}
\newcommand\nostarsection[1]
{\sectionprelude\origsection{#1}\sectionpostlude}
\newcommand\starsection[1]
{\sectionprelude\origsection*{#1}\sectionpostlude}
\newcommand\sectionprelude{\vspace{-3mm}}
\newcommand\sectionpostlude{\vspace{-2.5mm}}
\makeatother

\makeatletter
\let\origsubsection\subsection
\renewcommand\subsection{\@ifstar{\starsubsection}{\nostarsubsection}}
\newcommand\nostarsubsection[1]
{\subsectionprelude\origsubsection{#1}\subsectionpostlude}
\newcommand\starsubsection[1]
{\subsectionprelude\origsubsection*{#1}\subsectionpostlude}
\newcommand\subsectionprelude{\vspace{-2mm}}
\newcommand\subsectionpostlude{\vspace{-2.5mm}}
\makeatother

\graphicspath{ {figures/} }

\sethlcolor{tokbg}

\tcbset{
  decodedpanel/.style={
    enhanced jigsaw,
    breakable,
    boxrule=0pt,
    arc=3.2pt,
    boxsep=0pt,
    left=6pt,
    right=6pt,
    top=5pt,
    bottom=6pt,
    before skip=6pt,
    after skip=6pt,
    coltext=panelInk,
  },
}

\newtcolorbox{decodedpromptpanel}{
  decodedpanel,
  colback=panelPrompt,
  fontupper=\sffamily\small\linespread{1.15}\selectfont,
}
\newtcolorbox{decodedreasoningpanel}{
  decodedpanel,
  colback=panelReason,
  fontupper=\ttfamily\scriptsize\linespread{1.15}\selectfont,
  before upper={\raggedright\sloppy},
}
\newtcolorbox{decodedanswerpanel}{
  decodedpanel,
  colback=panelAnswer,
  fontupper=\sffamily\small\linespread{1.15}\selectfont,
}

\newtcblisting{extractiontemplate}{
  decodedpanel,
  colback=panelPrompt,
  listing only,
  listing options={style=extracttemplate},
}
\newtcblisting{extractioncall}{
  decodedpanel,
  colback=panelReason,
  listing only,
  listing options={style=extractcall},
}
\newtcblisting{judgepromptpanel}{
  decodedpanel,
  colback=panelPrompt,
  listing only,
  listing options={style=judgeprompt},
}

\usepackage{arydshln}
\makeatletter
\def\adl@drawiv#1#2#3{%
        \hskip.5\tabcolsep
        \xleaders#3{#2.5\@tempdimb #1{1}#2.5\@tempdimb}%
                #2\z@ plus1fil minus1fil\relax
        \hskip.5\tabcolsep}
\newcommand{\cdashlinelr}[1]{%
  \noalign{\vskip 2pt
           \global\let\@dashdrawstore\adl@draw
           \global\let\adl@draw\adl@drawiv}
  \cdashline{#1}[.4pt/2pt]
  \noalign{\global\let\adl@draw\@dashdrawstore
           \vskip 2pt}}
\makeatother

\usepackage{amsmath,amsfonts,bm}

\def\eqref#1{equation~\ref{#1}}

\def\1{\bm{1}}

\DeclareMathAlphabet{\mathsfit}{\encodingdefault}{\sfdefault}{m}{sl}
\SetMathAlphabet{\mathsfit}{bold}{\encodingdefault}{\sfdefault}{bx}{n}

\usepackage{subcaption}   

\providecommand{\figfont}{\rmfamily}
\usetikzlibrary{arrows.meta,calc,positioning}
\usepackage{microtype}
\makeatletter
\ifPDFTeX
  \microtypesetup{tracking=true}
\else\@ifpackagelater{microtype}{2024/12/12}
  {\microtypesetup{tracking=true}}
  {
   \DeclareRobustCommand{\textls}[2][100]{%
     {\addfontfeatures{LetterSpace=\the\numexpr#1/10\relax}#2}}}
\fi
\makeatother
\usetikzlibrary{arrows.meta,calc}
\pgfdeclarelayer{qexbg}\pgfsetlayers{qexbg,main}

\definecolor{qexInk}{HTML}{24272C}    \definecolor{qexMute}{HTML}{6E7480}   \definecolor{qexRule}{HTML}{DCDFE4}
\definecolor{qexImgF}{HTML}{DCEBF6}   \definecolor{qexImgE}{HTML}{9CC0DE}   \definecolor{qexImgT}{HTML}{2E6FA8}
\definecolor{qexCapF}{HTML}{FBEFD9}   \definecolor{qexCapE}{HTML}{E1C48A}   \definecolor{qexCapT}{HTML}{9C6A12}
\definecolor{qexCoral}{HTML}{E07B54}  \definecolor{qexCoralF}{HTML}{FCE9E1}
\definecolor{qexPhF}{HTML}{EEF0F3}    \definecolor{qexPhE}{HTML}{C7CBD1}

\pgfkeys{/qex/.is family, /qex,
  image/.store in=\qexImage,
  caption/.store in=\qexCaption,
  sdec/.store in=\qexSdec,
  shol/.store in=\qexShol,
  s/.store in=\qexS,
  highlight/.store in=\qexHl,
  mismatch/.store in=\qexMm,
  imglabel/.store in=\qexImgLabel,
}

\newcommand{\qualexample}[2][]{%
  \pgfqkeys{/qex}{image=, caption=, sdec=, shol=, s=, highlight=0, mismatch=0, imglabel=generated image}%
  \pgfqkeys{/qex}{#1}%
  \begin{tikzpicture}[
    x=1mm, y=1mm, font=\sffamily\fontsize{5.8}{6.6}\selectfont, text=qexInk,
    chip/.style={rounded corners=2.4pt, inner xsep=2.6pt, inner ysep=1.6pt, line width=0.4pt,
                 font=\sffamily\fontsize{5.4}{6.2}\selectfont},
    capchip/.style={chip, fill=qexCapF, draw=qexCapE, text=qexCapT, align=center},
    respill/.style={chip, fill=white, draw=qexPhE, text=qexMute},
    resours/.style={chip, fill=qexCoral, draw=qexCoral, text=white, font=\sffamily\bfseries\fontsize{5.6}{6.4}\selectfont},
    grp/.style={font=\sffamily\bfseries\fontsize{5}{5.6}\selectfont, inner sep=0pt},
    hd/.style={font=\sffamily\fontsize{5.6}{6.4}\selectfont, text=qexMute, inner sep=0pt},
    cell/.style={inner sep=0pt},
    hl/.style={text=qexCoral},
  ]
  \def\cPv{35.5}\def\cPvb{42.5}\def\cPt{51.5}\def\cPtb{58.5}\def\cA{68}\def\cW{75.5}
  \def\yGrp{-19.4}\def\yHd{-23.4}\def\yFirst{-28.4}\def\rowh{4.2}
  \ifx\qexImage\empty
    \node[rounded corners=2pt, draw=qexPhE, fill=qexPhF, line width=0.5pt, minimum width=16mm, minimum height=12mm,
          inner sep=0pt, text=qexMute] (img) at (8,-6) {image};
  \else
    \begin{scope}\clip[rounded corners=2pt] (0,-12) rectangle (16,0);
      \node[inner sep=0pt] at (8,-6) {\includegraphics[width=16mm,height=12mm]{\qexImage}};
    \end{scope}
    \node[rounded corners=2pt, draw=qexImgE, line width=0.5pt, minimum width=16mm, minimum height=12mm, inner sep=0pt] (img) at (8,-6) {};
  \fi
  \node[hd, anchor=north] at (8,-12.8) {\qexImgLabel};
  \node[capchip, anchor=west] at (19,-6) {``\qexCaption''};
  \node[grp, text=qexImgT]  at ({(\cPv+\cPvb)/2},\yGrp) {image};
  \node[grp, text=qexCapT]  at ({(\cPt+\cPtb)/2},\yGrp) {caption};
  \node[grp, text=qexCoral] at ({(\cA+\cW)/2},\yGrp)    {agree \& weigh};
  \node[hd] at (\cPv,\yHd)  {$p^v_k$};  \node[hd] at (\cPvb,\yHd) {$\bar p^v_k$};
  \node[hd] at (\cPt,\yHd)  {$p^t_k$};  \node[hd] at (\cPtb,\yHd) {$\bar p^t_k$};
  \node[hd] at (\cA,\yHd)   {$a_k$};    \node[hd] at (\cW,\yHd)   {$w_k$};
  \node[hd, anchor=west] at (1,\yHd) {question};
  \draw[qexRule, line width=0.4pt] (0,\yHd-2.2) -- (80,\yHd-2.2);
  \foreach \q/\pv/\pvb/\pt/\ptb/\a/\w [count=\k] in {#2} {
    \pgfmathsetmacro{\yy}{\yFirst-(\k-1)*\rowh}
    \xdef\qexN{\k}
    \gdef\qexIsHl{0}\foreach \h in \qexHl {\ifnum\h=\k\relax\gdef\qexIsHl{1}\fi}
    \gdef\qexIsMm{0}\foreach \h in \qexMm {\ifnum\h=\k\relax\gdef\qexIsMm{1}\fi}
    \ifnum\qexIsHl=1
      \node[cell, anchor=west, hl] at (1,\yy) {$q_{\k}$\ \q};
      \node[cell] at (\cPv,\yy) {\pv};  \node[cell] at (\cPvb,\yy) {\pvb};
      \node[cell, hl] at (\cPt,\yy) {\pt}; \node[cell, hl] at (\cPtb,\yy) {\ptb};
      \node[cell] at (\cA,\yy) {\a};    \node[cell, hl, font=\sffamily\bfseries\fontsize{5.8}{6.6}\selectfont] at (\cW,\yy) {\w};
    \else\ifnum\qexIsMm=1
      \node[cell, anchor=west, hl] at (1,\yy) {$q_{\k}$\ \q};
      \node[cell, hl] at (\cPv,\yy) {\pv}; \node[cell, hl] at (\cPvb,\yy) {\pvb};
      \node[cell, hl] at (\cPt,\yy) {\pt}; \node[cell, hl] at (\cPtb,\yy) {\ptb};
      \node[cell, hl, font=\sffamily\bfseries\fontsize{5.8}{6.6}\selectfont] at (\cA,\yy) {\a}; \node[cell] at (\cW,\yy) {\w};
    \else
      \node[cell, anchor=west] at (1,\yy) {$q_{\k}$\ \q};
      \node[cell] at (\cPv,\yy) {\pv};  \node[cell] at (\cPvb,\yy) {\pvb};
      \node[cell] at (\cPt,\yy) {\pt};  \node[cell] at (\cPtb,\yy) {\ptb};
      \node[cell] at (\cA,\yy) {\a};    \node[cell] at (\cW,\yy) {\w};
    \fi\fi
  }
  \pgfmathsetmacro{\yLast}{\yFirst-(\qexN-1)*\rowh}
  \begin{pgfonlayer}{qexbg}
    \fill[qexImgF, rounded corners=2pt]   (\cPv-4,\yGrp+2.2)  rectangle (\cPvb+4,\yLast-2.1);
    \fill[qexCapF, rounded corners=2pt]   (\cPt-4,\yGrp+2.2)  rectangle (\cPtb+4,\yLast-2.1);
    \fill[qexCoralF, rounded corners=2pt] (\cA-4.2,\yGrp+2.2) rectangle (\cW+4.2,\yLast-2.1);
  \end{pgfonlayer}
  \draw[qexRule, line width=0.4pt] (0,\yLast-2.6) -- (80,\yLast-2.6);
  \ifx\qexS\empty
    \node[resours, anchor=east] at (79.5,\yLast-5.4) {$S_{\mathrm{dec}} = \qexSdec$};
  \else
    \node[resours,  anchor=east] (rS)   at (79.5,\yLast-5.4) {$S = \qexS$};
    \node[respill, anchor=east] (rHol) at ([xshift=-1.5mm]rS.west) {$S_{\mathrm{hol}} = \qexShol$};
    \node[respill, anchor=east]        at ([xshift=-1.5mm]rHol.west) {$S_{\mathrm{dec}} = \qexSdec$};
  \fi
  \end{tikzpicture}}

\usepackage{pgfplots}
\usepgfplotslibrary{groupplots}
\usetikzlibrary{calc}
\pgfplotsset{compat=1.18}

\definecolor{cDSG}{HTML}{8A8A8A}  
\definecolor{cTIFA}{HTML}{5B8DBE}  
\definecolor{cVQA}{HTML}{3E9C88}  
\definecolor{cOurs}{HTML}{D9812E}  
\definecolor{bTIFA}{HTML}{3F6FAF}  
\definecolor{bRichHF}{HTML}{7FB3D5}  
\definecolor{bGenAIBench}{HTML}{3E9C88}  
\definecolor{bWinoground}{HTML}{B07AA1}  
\definecolor{bNegBench}{HTML}{D9812E}  
\definecolor{cFamGemmanovisenc}{HTML}{CC79A7}  
\definecolor{cFamInternVL}{HTML}{8C5A2B}  
\definecolor{cFamQwen}{HTML}{6A3D9A}  
\definecolor{cFamGemmadense}{HTML}{824E49}  
\definecolor{cFamGemmaembedlayer}{HTML}{DC9BD7}  
\definecolor{cWin}{HTML}{009E73}   
\definecolor{cInk}{HTML}{222222}
\definecolor{cMute}{HTML}{6E6E6E}
\definecolor{cRule}{HTML}{BDBDBD}

\pgfplotsset{
  resfig/.style={
    scale only axis, text=cInk,
    tick label style={font=\scriptsize, text=cMute}, label style={font=\footnotesize, text=cInk},
    title style={font=\footnotesize, at={(0.5,1)}, anchor=south, yshift=2pt},
    axis line style={draw=none}, tick style={draw=none},
    grid style={black!9, line width=0.4pt},
    scaled ticks=false, tick label style={/pgf/number format/fixed},
    every axis plot/.append style={line join=round, line cap=round},
  },
  pct/.style={yticklabel={\pgfmathprintnumber{\tick}\%}},
}

\tikzset{
  mDSG/.style={cDSG, mark=*, mark size=1.5pt, line width=1.0pt, mark options={solid, fill=white, draw=cDSG, line width=0.7pt}},
  mTIFA/.style={cTIFA, mark=*, mark size=1.5pt, line width=1.0pt, mark options={solid, fill=white, draw=cTIFA, line width=0.7pt}},
  mVQA/.style={cVQA, mark=*, mark size=1.5pt, line width=1.0pt, mark options={solid, fill=white, draw=cVQA, line width=0.7pt}},
  mOurs/.style={cOurs, mark=*, mark size=1.5pt, line width=1.3pt, mark options={solid, fill=cOurs, draw=cOurs, line width=0.7pt}},
  lg/.style={baseline={([yshift=-0.5ex]current bounding box.center)}},  
}

\newtcolorbox{promptbox}[1]{
  enhanced,
  title={#1},
  colframe=gray!50!black,       
  colbacktitle=gray!50!black,   
  coltitle=white,               
  fonttitle=\bfseries\sffamily\small,
  colback=white,                
  boxrule=0.8pt,                
  arc=3pt,                      
  left=8pt, right=8pt, top=8pt, bottom=8pt, 
  fontupper=\fontsize{8.5}{11}\ttfamily     
}

\newcommand{\model}{Depict}

\newif\ifswordsmalltables
\swordsmalltablestrue
\let\origresizebox\resizebox
\newsavebox{\swordtablebox}
\newcommand{\swordsmalltables}{%
  \ifswordsmalltables
    \let\small\footnotesize
    \renewcommand{\resizebox}[3]{%
      \sbox\swordtablebox{##3}%
      \ifdim\wd\swordtablebox>##1\relax
        \origresizebox{##1}{##2}{\usebox\swordtablebox}%
      \else
        \usebox\swordtablebox
      \fi}%
  \fi}
\AddToHook{env/table/begin}{\swordsmalltables}
\AddToHook{env/table*/begin}{\swordsmalltables}
\let\captionsmall\small
\DeclareCaptionFont{captionsmall}{\captionsmall}
\renewenvironment{abstract}
  {\vspace{0.22cm}%
   {\labelfont\fontsize{10.2}{12}\selectfont Abstract\par}%
   \vspace{0.04cm}%
   \begingroup\small}
  {\par\endgroup\vspace{0.35cm}}

\providecommand{\swordbodyfontvariant}{0}
\ifnum\swordbodyfontvariant=1\relax
  \let\paperromanfont\rmfamily
\fi

\providecommand{\swordtitlefontvariant}{0}
\ifnum\swordtitlefontvariant=1\relax
  \ifPDFTeX
    \newcommand{\swordtitleface}{\fontfamily{XCharter-TLF}\selectfont}
  \else
    \newfontfamily\swordtitleface{XCharter}[
      UprightFont    = XCharter-Roman,
      ItalicFont     = XCharter-Italic,
      BoldFont       = XCharter-Bold,
      BoldItalicFont = XCharter-BoldItalic
    ]
  \fi
\else
  \newcommand{\swordtitleface}{}
\fi

\providecommand{\swordtitleruleplacement}{1}
\providecommand{\swordtitlebrandvariant}{0}
\newcommand{\swordtitlelogo}{%
  \ifnum\swordtitlebrandvariant=1\relax
    \includegraphics[width=2.25cm]{assets/sword-logo-black.pdf}%
  \else
    \includegraphics[width=2.25cm]{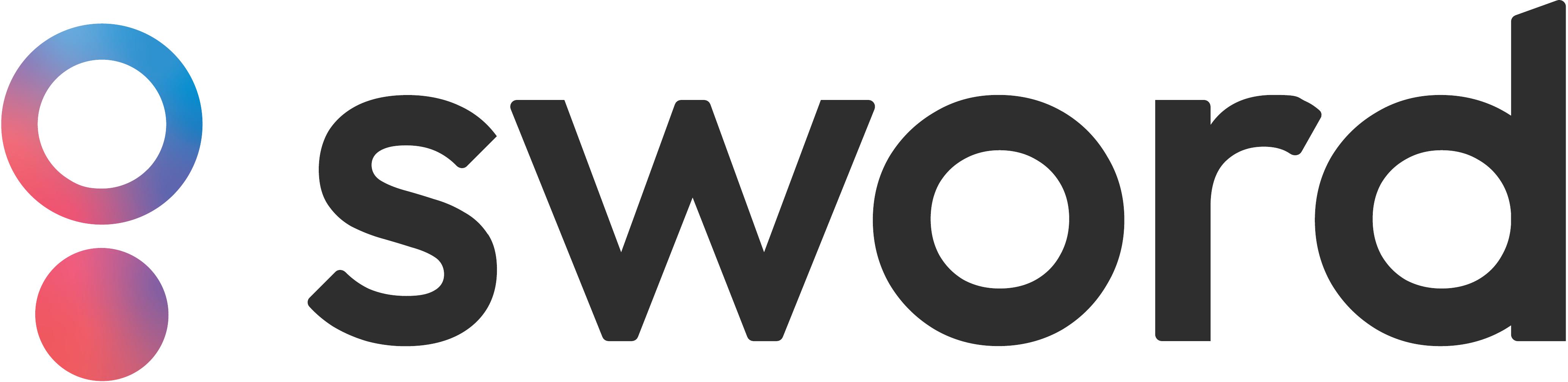}%
  \fi
}
\newcommand{\swordgradientrule}{%
  \begingroup
    \setlength{\parskip}{0pt}%
    \noindent
    \begin{tikzpicture}[baseline=0pt]
      \ifnum\swordtitlebrandvariant=0\relax
        \shade[
          shading=axis,
          left color=swordcoral,
          middle color=swordviolet,
          right color=swordblue,
          shading angle=90,
          rounded corners=0.7pt
        ] (0,0) rectangle (\linewidth,1.4pt);
      \else
        \fill[black,rounded corners=0.7pt]
          (0,0) rectangle (\linewidth,1.4pt);
      \fi
    \end{tikzpicture}\par
  \endgroup
}
\makeatletter
\renewcommand{\@maketitle}{%
  \newpage
  \null
  \vspace{-0.9cm}%
  \begingroup
    \setlength{\parindent}{0pt}%
    \raggedright
    \ifnum\swordtitleruleplacement=0\relax
      \swordgradientrule
      \vspace{0.22cm}%
    \fi
    \swordtitlelogo\par
    \ifnum\swordtitleruleplacement=1\relax
      \vspace{-0.08cm}%
      \swordgradientrule
      \vspace{0.22cm}%
    \else
      \vspace{0.25cm}%
    \fi
    {\swordtitleface\LARGE\bfseries\@title\par}
    \vspace{0.22cm}%
    {\@author\par}
  \endgroup
}
\makeatother

\begin{document}
\maketitle
\begin{abstract}
Image-text alignment is a core problem in computer vision with applications in caption evaluation, hallucination detection, data curation, and the benchmarking of text-to-image (T2I) generators. As T2I models improve, benchmarking has become demanding, requiring metrics capable of finding a series of issues like missing objects, swapped attributes, miscounts, and ignored negations. Recent work addresses this by fine-tuning evaluators on preference data or by prompting a vision-language model, either holistically with the caption or with decomposed verification questions. However, existing approaches fall short: fine-tuned metrics remain bound to one backbone and training distribution; holistic metrics miss fine-grained details; and decomposed metrics rely on a fixed-YES assumption that penalizes faithful images whenever that assumption fails. In contrast, we propose \textsc{\model}, a training-free metric that replaces fixed reference answers with expected agreement between image-based and caption-only answers, weighting questions by how decisively the caption determines them. By replacing fixed references, our agreement rule increases negation accuracy from 19\% to 88\%. To recover the context lost during decomposition, \textsc{\model} merges this agreement score with a holistic score.
We evaluate \textsc{\model} on five benchmarks and eleven backbones from three model families and find that it surpasses all training-free metrics and exceeds fine-tuned evaluators on two out of three human-correlation benchmarks.\footnote{Code will be available for research purposes.}
\end{abstract}

\section{Introduction}
\label{sec:intro}

Image-text alignment evaluation is fundamental across vision-language tasks, including caption scoring~\citep{Hessel_2021_clipscore}, vision-language models (VLM) hallucination detection~\citep{kogilathota2026halp}, data filtering~\citep{schuhmann2021laion,gadre2023datacomp}, and, increasingly, evaluating text-to-image (T2I) generators~\citep{Kamath_2025_geneval2,noisyclip} and their reward signals~\citep{Xu_2023_imagereward,black2024trainingddpo}. As T2I models improve, the question has shifted from whether they can generate an image that matches a prompt to whether they can render every aspect of it faithfully, without dropping an object, swapping an attribute, miscounting, or ignoring a negation~\citep{flux,chefer2026self,qwenimage}. An alignment metric must therefore correlate with human judgments on open-ended text, including these fine-grained errors~\citep{ku2024imagenhub}.

Two families of evaluators have emerged. Fine-tuned metrics learn a scoring function from preference data~\citep{Xu_2023_imagereward,Kirstain_2023_pickapic,Wu_2023_human,Marjit_2026_dyneval}, while training-free ones prompt an off-the-shelf VLM~\citep{vqascore,Hu_2023_tifa,Cho_2024_davidsonian}. We focus on the training-free family due to a fundamental structural advantage: while fine-tuned evaluators remain bound to a specific architecture and the static distribution of their training data, training-free metrics naturally scale alongside foundation models, inheriting backbone advancements without task-specific re-training. To evaluate how effectively these metrics leverage underlying backbone improvements, we conduct what is, to our knowledge, the first study assessing them across eleven backbones from three model families and five benchmarks.

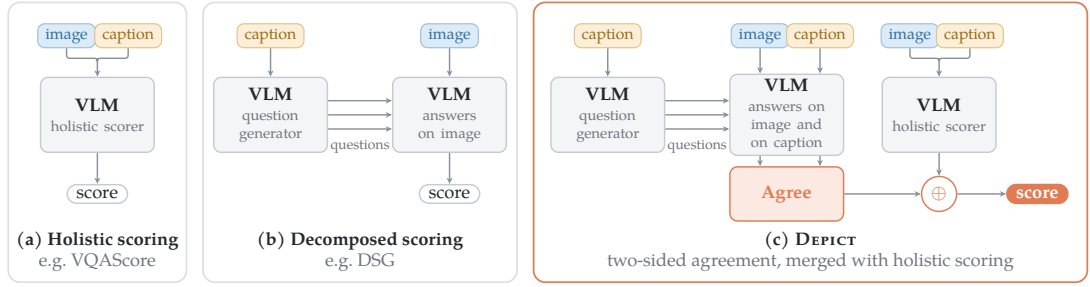
\begin{figure}[t]
    \centering
    \resizebox{0.9\textwidth}{!}{
\definecolor{ink}{HTML}{24272C}%
\definecolor{mute}{HTML}{6E7480}%
\definecolor{arrowc}{HTML}{8C929B}%
\definecolor{frame}{HTML}{DCDFE4}%
\definecolor{cardF}{HTML}{F4F5F7}%
\definecolor{cardE}{HTML}{C7CBD1}%
\definecolor{imgF}{HTML}{DCEBF6}%
\definecolor{imgE}{HTML}{9CC0DE}%
\definecolor{imgT}{HTML}{2E6FA8}%
\definecolor{capF}{HTML}{FBEFD9}%
\definecolor{capE}{HTML}{E1C48A}%
\definecolor{capT}{HTML}{9C6A12}%
\definecolor{coral}{HTML}{E07B54}%
\definecolor{coralF}{HTML}{FCE9E1}%
\begin{tikzpicture}[
  x=1mm, y=1mm, font=\figfont\fontsize{6.2}{7.2}\selectfont, text=ink,
  >={Stealth[length=2.2pt,width=2pt]},
  chip/.style={rounded corners=2.4pt, inner xsep=2.6pt, inner ysep=1.5pt, line width=0.4pt,
               font=\figfont\fontsize{5.4}{6.2}\selectfont},
  imgchip/.style={chip, fill=imgF, draw=imgE, text=imgT},
  capchip/.style={chip, fill=capF, draw=capE, text=capT},
  card/.style={rounded corners=2.6pt, draw=cardE, fill=cardF, line width=0.5pt,
               text width=12.6mm, minimum height=9mm, inner sep=1.5pt, align=center,
               font=\figfont\fontsize{6.2}{6.2}\selectfont},
  ours/.style={draw=coral, fill=coralF, text=coral},
  pill/.style={rounded corners=3.2pt, draw=cardE, fill=white, line width=0.5pt,
               inner xsep=3.2pt, inner ysep=1.8pt},
  pillours/.style={pill, draw=coral, fill=coral, text=white, font=\figfont\bfseries\fontsize{6.2}{7.2}\selectfont},
  plus/.style={circle, draw=coral, fill=white, line width=0.5pt, inner sep=0pt, minimum size=4.2mm, text=coral},
  h/.style={font=\figfont\bfseries\fontsize{6.2}{7.2}\selectfont},
  s/.style={font=\figfont\fontsize{5.2}{6}\selectfont, text=mute},
  arr/.style={->, draw=arrowc, line width=0.5pt, rounded corners=1pt},
  panel/.style={rounded corners=4pt, draw=frame, line width=0.6pt},
]
\newcommand{\role}[2]{{\bfseries #1}\\[0.15ex]{\fontsize{5}{5.6}\selectfont\color{mute}#2}}
\newcommand{\qarrows}[3]{
  \foreach \dy in {1.7, 0, -1.7}
    \draw[arr] ($(#1.east)+(0,\dy)$) -- ($(#2.west)+(0,\dy)$);
  \node[s, font=\figfont\fontsize{4.8}{5.4}\selectfont] at (#3, -13.3) {questions};}

\node[imgchip] (a-img) at (8,0) {image};
\node[capchip] (a-cap) at (15.4,0) {caption};
\node[card] (a-vlm) at (11.7,-9.5) {\role{VLM}{holistic scorer}};
\node[pill] (a-sc) at (11.7,-19) {score};
\coordinate (a-j) at ($(a-vlm.north)+(0,2.2)$);
\draw[arr,-] (a-img.south) -- ++(0,-1.2) -| (a-j);
\draw[arr,-] (a-cap.south) -- ++(0,-1.2) -| (a-j);
\draw[arr] (a-j) -- (a-vlm.north);
\draw[arr] (a-vlm.south) -- (a-sc.north);
\draw[panel] (1,4) rectangle (22.4,-29.6);
\node[h, align=center] at (11.7,-25.9) {(a) Holistic scoring\\{\mdseries\color{mute}e.g.\ VQAScore}};

\node[capchip] (b-cap) at (32.5,0) {caption};
\node[imgchip] (b-img) at (54,0) {image};
\node[card] (b-qg) at (32.5,-9.5) {\role{VLM}{question generator}};
\node[card] (b-vlm) at (54,-9.5) {\role{VLM}{answers on image}};
\node[pill] (b-sc) at (54,-19) {score};
\draw[arr] (b-cap.south) -- (b-qg.north);
\draw[arr] (b-img.south) -- (b-vlm.north);
\qarrows{b-qg}{b-vlm}{43.25}
\draw[arr] (b-vlm.south) -- (b-sc.north);
\draw[panel] (24.4,4) rectangle (62.1,-29.6);
\node[h, align=center] at (43.25,-25.9) {(b) Decomposed scoring\\{\mdseries\color{mute}e.g.\ DSG}};

\node[capchip] (c-cap) at (73,0) {caption};
\node[imgchip] (c-img1) at (91.3,0) {image};
\node[capchip] (c-cap2) at (98.5,0) {caption};
\node[imgchip] (c-img2) at (109.3,0) {image};
\node[capchip] (c-cap3) at (116.5,0) {caption};
\node[card] (c-qg) at (73,-9.5) {\role{VLM}{question generator}};
\node[card] (c-ver) at (94.5,-9.5) {\role{VLM}{answers on image and on caption}};
\node[card] (c-hol) at (112.8,-9.5) {\role{VLM}{holistic scorer}};
\node[card, ours, minimum height=6.5mm, font=\figfont\bfseries\fontsize{6.2}{6.2}\selectfont] (c-agr) at (94.5,-19) {Agree};
\node[plus] (c-plus) at (112.8,-19) {$\oplus$};
\node[pillours] (c-sc) at (124.6,-19) {score};
\draw[arr] (c-cap.south) -- (c-qg.north);
\draw[arr] (c-img1.south) -- (c-img1.south |- c-ver.north);
\draw[arr] (c-cap2.south) -- (c-cap2.south |- c-ver.north);
\coordinate (c-j) at ($(c-hol.north)+(0,2.2)$);
\draw[arr,-] (c-img2.south) -- ++(0,-1.2) -| (c-j);
\draw[arr,-] (c-cap3.south) -- ++(0,-1.2) -| (c-j);
\draw[arr] (c-j) -- (c-hol.north);
\qarrows{c-qg}{c-ver}{83.75}
\draw[arr] (c-img1.south |- c-ver.south) -- (c-img1.south |- c-agr.north);
\draw[arr] (c-cap2.south |- c-ver.south) -- (c-cap2.south |- c-agr.north);
\draw[arr] (c-hol.south) -- (c-plus.north);
\draw[arr] (c-agr.east) -- (c-plus.west);
\draw[arr] (c-plus.east) -- (c-sc.west);
\draw[panel, draw=coral] (64.1,4) rectangle (130.6,-29.6);
\node[h, align=center] at (97.35,-25.9) {(c) \textsc{Depict}\\{\mdseries\color{mute}two-sided agreement, merged with holistic scoring}};
\end{tikzpicture}}
    \caption{\textbf{Text-to-image scoring paradigms.} (a)~\emph{Holistic}: directly scores the image-caption pair. (b)~\emph{Decomposed}: answers caption-generated questions on the image. (c)~\textsc{\model}: checks agreement between image- and caption-based answers, then merges it with a holistic score.}
    \label{fig:teaser}
\end{figure}

Within the training-free family, two approaches have developed separately. \emph{Holistic} scoring started with global embedding similarity between separately encoded text and image~\citep{Hessel_2021_clipscore} and newer approaches ask VLMs a single question about the whole caption and read the probability of \textsc{yes}~(\cite{vqascore};  Figure~\ref{fig:teaser}a). \emph{Decomposed} scoring methods generate a set of verification questions from the caption, answer each against the image, and aggregate the results to obtain a final score~(\cite{Hu_2023_tifa,Cho_2024_davidsonian,Kamath_2025_geneval2}; Figure~\ref{fig:teaser}b). Holistic scoring preserves macro context and decomposed metrics offer fine-grained interpretability, but neither approach dominates across benchmarks. Decomposed metrics also share a flaw in how each answer is scored. The original TIFA scheme graded each question by string-matching the VLM's answer against a label derived from the caption, such as objects, counts, or actions~\citep{Hu_2023_tifa}. Later methods simplified this by assuming a constant \textsc{yes} target, scored either by hard string matches~\citep{Cho_2024_davidsonian} or their probabilities~\citep{Kamath_2025_geneval2}. We call this the \emph{fixed-\textsc{yes} assumption}: the expected answer is fixed before looking at the image, so whenever it is wrong, the metric fails with it.

We propose \textsc{\model}, a training-free alignment metric (Figure~\ref{fig:teaser}c). Given a caption, \textsc{\model} decomposes the description into a set of visual assertions without requiring ground-truth labels. A VLM evaluates each assertion twice, once against the image and once against the caption alone, assigning a verdict to each. The score reflects the agreement between the two channels, which is high whenever both favor the same truth value, regardless of polarity. We further find that the holistic and decomposed paradigms are complementary rather than competing and \textsc{\model} is the first metric to merge them. Across every human-judgment benchmark we test, the merged score correlates better with human ratings than either approach alone.

Our contributions are:

\begin{itemize}
    \item \textbf{Diagnosis of the fixed-\textsc{yes} flaw.} We uncover a fundamental structural flaw in current decomposed metrics, the fixed-\textsc{yes} assumption, which causes their accuracy on negated prompts to drop below chance.

    \item \textbf{Agreement-based scoring.} We score each question by the agreement between a text-only and a visual answer, which removes the assumption and improves both negation accuracy and correlation with human judgements.

    \item \textbf{Systematic benchmark study.} We present, to our knowledge, the first comparison of training-free T2I alignment metrics across backbones from several model families and multiple benchmarks, with paired confidence intervals for every difference.

    \item \textbf{\textsc{\model}.} We merge agreement-based decomposed scoring with holistic scoring in a training-free metric. It exceeds the strongest fine-tuned evaluator on its own backbone on most human-judgment benchmarks and consistently matches or beats training-free baselines across backbones and benchmarks.
\end{itemize}

\section{Related Work}
\label{sec:related_work}

\paragraph{Embedding-based metrics.} 
Early alignment metrics like CLIPScore~\citep{Hessel_2021_clipscore} and BLIP2Score~\citep{li2023blip2} rely on global image-text similarity. Because these models are pre-trained for coarse caption matching rather than fine-grained relational reasoning, they behave as bag-of-words matchers~\citep{Yuksekgonul0KJ023_bag_of_words,koishigarina2026clip_bag_of_words} that struggle with compositionality and correlate weakly with human judgments~\citep{Hu_2023_tifa,Cho_2024_davidsonian}.

\paragraph{Fine-tuned metrics.} The initial approaches take the encoders above and align their outputs with human preference data~\citep{Xu_2023_imagereward,Kirstain_2023_pickapic,Wu_2023_human}, while newer evaluators start from a VLM and learn from human ratings or from a larger teacher model~\citep{tu-etal-2025-automatic,Wang_2026_unigenbench,Marjit_2026_dyneval}. In both cases, the learned score is bound to its backbone and its training data. DynEval~\citep{Marjit_2026_dyneval}, the strongest of these, fine-tunes a decomposition pipeline on Qwen3-VL~\citep{yang2025qwen3} and serves as our matched-backbone reference.

\paragraph{Training-free metrics.} These metrics query an off-the-shelf VLM~\citep{yang2025qwen3,team2026gemma}, so any stronger backbone can be swapped in at no cost. Early approaches decomposed the caption into a question per assertion~\citep{vq2,singh2023divide} and checked the image's answer against a reference (a caption-derived label in TIFA~\citep{Hu_2023_tifa} or a fixed \textsc{yes} in DSG~\citep{Cho_2024_davidsonian}). Soft scoring then replaced exact matching. VQAScore~\citep{vqascore} evaluated the entire caption by reading the probability of a single \textsc{yes}, capturing overall context but remaining blind to fine-grained details. Meanwhile, Gecko~\citep{wiles2024gecko} and Soft-TIFA~\citep{Kamath_2025_geneval2} read per-question answer probabilities, making decomposed scores continuous while retaining the fixed reference. \textsc{\model} replaces this static reference with the agreement between a text and a visual answer and is the first to merge the decomposed and holistic approaches.
\section{Measuring Alignment Between Text and Vision}
\label{sec:method}

\begin{figure}[t]
    \centering
    \resizebox{0.9\textwidth}{!}{\input{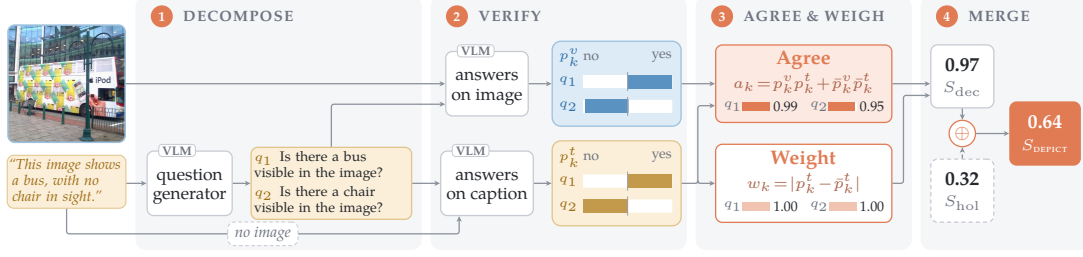}}
    \caption{Overview of \textsc{\model}. \textbf{(1)} A VLM converts the caption into yes/no questions $q_k$. \textbf{(2)} A frozen VLM answers each question on the image ($p^v_k$) and on the caption alone ($p^t_k$). \textbf{(3)} Questions are scored by answer agreement $a_k$ (rewarding matching \emph{no} answers) and weighted by commitment $w_k$. \textbf{(4)} The weighted decomposed score $S_{\mathrm{dec}}$ is averaged with holistic score $S_{\mathrm{hol}}$.}
    \label{fig:method}
\end{figure}

We present \textsc{\model}, a training-free metric that scores an image $I$ against a caption $t$. Rather than scoring each question against a fixed answer as prior decomposed metrics do~\citep{Hu_2023_tifa,Cho_2024_davidsonian}, \textsc{\model} treats the answer $y \in \{\textsc{yes}, \textsc{no}\}$ as a latent variable. It generates questions directly from the caption, answers each from the image and from the caption alone, and scores a question by the probability that the two answers coincide, marginalized over both outcomes. Questions are then weighted by how decisively the caption settles them, and the resulting decomposed score is merged with a holistic one (Figure~\ref{fig:method}).

\paragraph{Generating questions from the caption.}
DSG~\citep{Cho_2024_davidsonian} first parses $t$ into semantic tuples and then converts each tuple into a question. We skip the intermediate parsing step by prompting a language model to produce $N$ yes/no questions $\{q_k\}_{k=1}^{N}$ directly from $t$ , each targeting a single verifiable detail and answerable from the image alone. Unlike DSG, the generator sees the whole caption, so questions do not inherit possible parsing errors and retain the context that tuple extraction discards, making post-hoc filtering and a dependency graph unnecessary. We ablate this in Section~\ref{sec:ablations}.

\paragraph{Continuous soft scoring.}
Decomposed metrics such as TIFA~\citep{Hu_2023_tifa} and DSG~\citep{Cho_2024_davidsonian} score each question $q_k$ against a reference $y_k^*$ using a hard indicator function $a_k^{\text{hard}} = \mathbb{I}\left(\text{VQA}(I, q_k) = y_k^*\right)$. Following soft approaches~\citep{vqascore,Kamath_2025_geneval2}, we read the softmax probability of the tokens, $p^v_k = p(\textsc{yes} \mid I, q_k)$, rather than decoding a discrete answer. Probabilities are not renormalized over the \textsc{yes}/\textsc{no} pair, so $p(\textsc{yes}) + p(\textsc{no}) \le 1$ and mass placed elsewhere lowers the score, penalizing off-target outputs without forcing a binary choice.

\paragraph{From a fixed reference to expected agreement.}
Most decomposed evaluators score under a fixed-reference assumption, setting $y_k^{*}=\textsc{yes}$ for every question~\citep{Cho_2024_davidsonian, Kamath_2025_geneval2}. We relax this assumption progressively with the following scoring rules $a_k$:
\begin{align}
\text{Faith:}\quad   & a_k = p^v_k \label{eq:faith}\\
\text{Confirm:}\quad & a_k = p^v_k\,p^t_k \label{eq:confirm}\\
\text{Agree:}\quad   & a_k = p^v_k\,p^t_k + \bar p^v_k\,\bar p^t_k \label{eq:agree}.
\end{align}

Current soft metrics~\citep{vqascore,Kamath_2025_geneval2} rely on \emph{Faith}, measuring only the visual channel's probability of answering \textsc{yes} as in~\eqref{eq:faith}. \emph{Confirm} replaces the fixed reference with a soft text channel, $p^t_k = p(\textsc{yes} \mid t, q_k)$, in which the model answers the same question from the caption instead of the image (Eq.~\ref{eq:confirm}). This is the continuous version of a text-derived reference $y_k^*$ where the visual score is weighted by how strongly the caption entails \textsc{yes}~\citep{Hu_2023_tifa}. However, because \emph{Confirm} only rewards positive agreement, it fails whenever the correct answer is \textsc{no}. \emph{Agree} resolves this by incorporating \textsc{no} probabilities ($\bar p^v_k, \bar p^t_k$) and marginalizing over the latent answer as $a_k = \sum_{y} p(y \mid t, q_k)\, p(y \mid I, q_k) = p^t_k p^v_k + \bar p^t_k \bar p^v_k$ (Eq.~\ref{eq:agree}). It is symmetric across modalities and generalizes prior rules, reducing to hard discrete matching for one-hot outputs, to \emph{Faith} when $p^t_k{=}1$, and to \emph{Confirm} when omitting the \textsc{no} term. 

\paragraph{Commitment Weighting.}
The text channel's answers $p^t_k$ and $\bar p^t_k$ show how confidently the model can answer $q_k$ from the caption alone. When they are close, the model is uncertain and the question says little about alignment, whether because it is ambiguous or off-target. We therefore weight each question by the text channel's commitment: 
\begin{equation} 
w_k = \left| p^t_k - \bar p^t_k \right| , \label{eq:cw} 
\end{equation} 
so that questions answered decisively with the caption context dominate the score while uncertain ones vanish. This serves as a soft quality filter, eliminating the need for post-hoc pruning (we ablate this mechanism in Section~\ref{sec:ablations}).

\paragraph{Decomposed Score Aggregation.}
The final decomposed alignment score is the commitment-weighted mean of the per-question scores $a_k$ across all $N$ questions: \begin{equation} S_{\text{dec}}(I, t) = \frac{\sum_{k=1}^{N} w_k\, a_k}{\sum_{k=1}^{N} w_k + \epsilon}, \label{eq:aggregation} \end{equation} where $a_k$ is the per-question score of Eqs.~\ref{eq:faith}--\ref{eq:agree} and $\epsilon$ prevents division by zero.

\paragraph{Holistic Answering and Merge.}
\label{subsec:convex_merge}
Decomposed questions capture fine-grained visual compositional logic but occasionally lose global context, whereas holistic evaluation captures macro-level visual alignment but misses subtle compositional failures. Following VQAScore~\citep{vqascore}, we compute a holistic alignment score $S_{\text{hol}}(I, t)$ by querying the VLM with the entire caption $t$ directly:
\begin{equation}
S_{\text{hol}}(I, t) = p( \textsc{yes} \mid I, t).
\end{equation}
The \textsc{\model} score is the convex combination of the decomposed $S_{\text{dec}}(I, t)$ and the holistic $S_{\text{hol}}(I, t)$:
\begin{equation}
S_{\textsc{\model}}(I, t) = (1 - \lambda) \, S_{\text{dec}}(I, t) + \lambda \, S_{\text{hol}}(I, t) ,
\label{eq:merge}
\end{equation}
where $\lambda \in [0, 1]$ balances fine-grained compositional detail with macro-level relational context.
\section{Experimental Setup}
\label{sec:experimental_setup}

\paragraph{Datasets, metrics, and statistics.}
\label{sec:datasets}
\label{sec:stats}
To measure the correlation with human judgements, we use GenAI-Bench~\citep{baiqi-li2024genaibench}, TIFA160~\citep{Hu_2023_tifa}, and RichHF-18K~\citep{richhf} \textit{misalignment\_score}, reporting Spearman Rank Correlation
Coefficient (SRCC) on raw scores and Pearson Linear Correlation Coefficient (PLCC) following~\cite{Han_2024_evalmuse40k}. For diagnostics, we use Winoground~\citep{thrush2022winoground}, reporting text, image, and group accuracy, and the COCO-MCQ split of NegBench~\citep{alhamoud2025vision_negbench}. Score differences ($\Delta$) between metrics is paired, with BCa 95\% intervals over $B=10{,}000$ bootstrap replicates resampled by item. Dataset details are presented in Appendix~\ref{ann:datasets}.

\paragraph{Baselines.}
\label{sec:baselines}
We compare against three groups of metrics. \emph{1) Training-free, embedding-similarity} metrics score the cosine similarity of caption and image embeddings from a contrastive encoder: CLIPScore~\citep{Hessel_2021_clipscore} and BLIPv2Score~\citep{li2023blip2}. \emph{2) Training-free, VLM-based} metrics prompt an off-the-shelf VLM and run on any backbone; these are the focus of our analysis, and we take one representative per cell of the decomposition~$\times$~granularity grid of Section~\ref{sec:paired_comparisons}: DSG~\citep{Cho_2024_davidsonian} (decomposed, discrete), Soft-TIFA~\citep{Kamath_2025_geneval2} (decomposed, continuous; the Faith rule of Eq.~\ref{eq:faith}), and VQAScore~\citep{vqascore} (holistic, continuous). \emph{3) Fine-tuned evaluators} learn a scoring function from preference data or teacher annotations and are tied to their training backbone: ImageReward~\citep{Xu_2023_imagereward}, PickScore~\citep{Kirstain_2023_pickapic}, UniGenBench++~\citep{Wang_2026_unigenbench}, LongT2IBench~\citep{Yang_2025_longt2ibench}, T2I-Eval~\citep{tu-etal-2025-automatic}, FGA-BLIP2~\citep{Han_2024_evalmuse40k}, and DynEval~\citep{Marjit_2026_dyneval}.

\paragraph{Backbones.}
\label{sec:backbones}
To decouple the scoring procedure from visual reasoning capability, all training-free methods are run on eleven backbones varying along four axes: 1)~\emph{family}, via Qwen3.5~\citep{yang2025qwen3}, InternVL3.5~\citep{wang2025internvl3}, and Gemma-4~\citep{team2026gemma}; \emph{2) scale}, via two size ladders (Qwen3.5 4B/9B/27B, InternVL3.5 4B/8B/14B); \emph{3) generation}, via Qwen3.5/3.6/3.8 at a fixed 27B; and \emph{4) architecture}, via the Gemma-4 variants (per-layer embedding E4B, unified multimodal 12B, dense 31B). Qwen3-VL-4B provides a matched comparison against DynEval.

\paragraph{Implementation Details.}
\label{sec:impl}
Question generation and scoring use the same backbone, so every row of Tables~\ref{tab:main} and~\ref{tab:anchor} uses a single model; prompts are in Appendix~\ref{ann:prompts}. The generator has no question budget and produces as many questions as the caption requires, so $N$ varies per caption (statistics in Appendix~\ref{ann:datasets}). For each question we obtain the visual answer $p^{v}$ from the image-question pair and the textual answer $p^{t}$ from the caption-question pair with the image removed; answer probabilities are full-vocabulary softmax masses summed over the variants of each answer token (e.g.\ \texttt{yes}, \texttt{Yes}, \texttt{\textvisiblespace yes}). We use commitment weighting, no pruning, and arithmetic aggregation, with $\epsilon=10^{-3}$ in~\eqref{eq:aggregation} and $\lambda=0.5$ in~\eqref{eq:merge} fixed a priori and not tuned on any benchmark (sensitivity in Appendix~\ref{ann:lambda}). All training-free methods see identical images, prompts, and decoding settings: VQAScore uses its original prompt, DSG runs its published pipeline with only the VLM swapped, and Soft-TIFA-AM, which specifies only a scoring rule (an unweighted mean of $p^{v}$ against a fixed \textsc{yes}), is applied to our question sets, where it coincides with the unweighted Faith row of Table~\ref{tab:ablations}. Computational pass counts and latency analysis are detailed in Appendix~\ref{ann:cost}.

\begin{table}[t]
\centering
\small  
\setlength{\tabcolsep}{5.3pt}
\begin{tabular}{llcccccc}
\toprule
\multirow{2}{*}{\textbf{Method}} & \multirow{2}{*}{\textbf{Backbone}} & \multicolumn{2}{c}{\textbf{GenAI-Bench}} & \multicolumn{2}{c}{\textbf{TIFA160}} & \multicolumn{2}{c}{\textbf{RichHF}} \\ \cmidrule(lr){3-4}\cmidrule(lr){5-6}\cmidrule(lr){7-8}
 &  & SRCC & PLCC & SRCC & PLCC & SRCC & PLCC \\ \midrule
\multicolumn{8}{l}{\textit{Fine-tuned evaluators}} \\
ImageReward$^\dagger$ & BLIP & 0.340 & 0.379 & 0.621 & 0.634 & 0.275 & 0.329 \\
PickScore$^\dagger$ & CLIP-H & 0.354 & 0.363 & 0.428 & 0.434 & 0.392 & 0.413 \\
UniGenBench++$^\dagger$ & Qwen2.5-VL-72B & 0.461 & 0.444 & 0.608 & 0.590 & 0.533 & 0.517 \\
LongT2IBench$^\dagger$ & Qwen2.5-VL-7B & 0.375 & 0.376 & 0.678 & 0.692 & 0.517 & 0.516 \\
T2I-Eval$^\dagger$ & MiniCPM-V & 0.500 & 0.477 & 0.601 & 0.625 & 0.503 & 0.499 \\
FGA-BLIP2 (os)$^\dagger$ & BLIP-2 & 0.564 & 0.567 & 0.760 & 0.744 & 0.512 & 0.546 \\
DynEval-4B$^\dagger$ & Qwen3-VL-4B & 0.595 & 0.582 & \textbf{0.802} & \textbf{0.803} & 0.546 & 0.569 \\
\midrule
\multicolumn{8}{l}{\textit{Training-free, embedding similarity}} \\
CLIPScore$^\dagger$ & CLIP & 0.168 & 0.203 & 0.300 & 0.309 & 0.057 & 0.302 \\
BLIPv2Score$^\dagger$ & BLIP-2 & 0.273 & 0.298 & 0.429 & 0.454 & 0.143 & 0.310 \\
\midrule
\multicolumn{8}{l}{\textit{Training-free, VLM-based, original backbone}} \\
VQAScore & CLIP-FlanT5 & 0.553 & 0.518 & 0.695 & 0.659 & 0.483 & 0.409 \\
Soft-TIFA-AM$^\dagger$ & Qwen3-VL-8B & 0.482 & 0.457 & 0.682 & 0.673 & 0.631 & 0.602 \\
\midrule
\multicolumn{8}{l}{\textit{Training-free, VLM-based, current backbone}} \\
DSG          & Qwen3-VL-4B & 0.437 & 0.441 & 0.610 & 0.613 & 0.489 & 0.472 \\
Soft-TIFA-AM & Qwen3-VL-4B & 0.418 & 0.412 & 0.667 & 0.666 & 0.528 & 0.534 \\
VQAScore     & Qwen3-VL-4B & 0.628 & 0.594 & 0.658 & 0.591 & \underline{0.607} & 0.548 \\
\textsc{\model} & Qwen3-VL-4B & \underline{0.630} & \underline{0.627} & 0.691 & 0.675 & \underline{0.607} & \underline{0.597} \\
\addlinespace[2pt]
\textsc{\model} & Qwen3.5-27B & \textbf{0.720} & \textbf{0.725} & \underline{0.780} & \underline{0.767} & \textbf{0.667} & \textbf{0.664} \\
\bottomrule
\end{tabular}
\caption{\textbf{Human correlation} on GenAI-Bench, TIFA160, and RichHF-18K. \textbf{Bold} marks best value per column and \underline{underline} second best. $^\dagger$: cited from publications under matching protocol; others re-run. Qwen3-VL-4B matches top baseline backbone; final row uses our largest backbone.}
\label{tab:main}
\end{table}

\section{Results and Discussion}
\label{sec:results_and_discussion}

We evaluate \textsc{\model} against fine-tuned and training-free baselines across human alignment, compositional discrimination, and explicit negation. Across five benchmarks and eleven backbones, we show that expected agreement and term merging match or exceed fine-tuned evaluators, hold across backbones, and avoid failures of fixed-\textsc{yes} metrics. Qualitative evaluation is provided in Appendix~\ref{ann:qualitative}.

\subsection{Does \textsc{\model} compete with fine-tuned evaluators?}

Table~\ref{tab:main} compares \textsc{\model} with fine-tuned evaluators, and with training-free metrics on both their original and current backbones, including Qwen3-VL-4B, the backbone of the strongest fine-tuned evaluator, which allows a matched comparison.

\paragraph{Competitive with fine-tuned evaluators.} 
\textsc{\model} is best or second best across all datasets and metrics of Table~\ref{tab:main}. When evaluated on the same Qwen3-VL-4B backbone as DynEval-4B, \textsc{\model} surpasses the strongest fine-tuned baseline on GenAI-Bench and RichHF, trailing only on TIFA160. Furthermore, on this shared backbone, \textsc{\model} matches or exceeds VQAScore (the second-best training-free approach) in SRCC across all three benchmarks while strictly outperforming it in PLCC.

\paragraph{Training-free metrics track the backbone.}
Swapping VQAScore's 2024 backbone for Qwen3-VL-4B, with no other change, raises its SRCC by $0.075$ on GenAI-Bench and $0.124$ on RichHF, enough to surpass every fine-tuned evaluator in the table on those two benchmarks, including one trained on a 72B model. Moving \textsc{\model} from Qwen3-VL-4B to Qwen3.5-27B adds a further $0.06$--$0.10$ in every column and yields the best score on GenAI-Bench and RichHF.

\subsection{Where do the gains come from?}
\label{sec:paired_comparisons}
To separate procedure from backbone capability, Table~\ref{tab:anchor} fixes the backbone at Qwen3.5-27B, so training-free methods differ only in how they score. The rows map onto the components of Section~\ref{sec:method}: DSG uses its own tuple-based questions with hard scoring, Soft-TIFA-AM scores our questions with \emph{Faith}, VQAScore is the holistic term, and \textsc{\model}'s decomposed term scores the same questions with \emph{Agree} before being merged with holistic term.

\paragraph{The baselines split by benchmark.}
Comparing training-free baselines on a shared backbone reveals neither paradigm dominates. Soft-TIFA leads on TIFA160, whereas VQAScore leads on GenAI-Bench and RichHF (Table~\ref{tab:anchor}). TIFA160 favors decomposition in general, with decomposed metrics scoring $0.09\text{-}0.22$ higher on TIFA160 than on GenAI-Bench (against $0.04$ for VQAScore) which is a pattern consistent across every decomposed evaluator in Table~\ref{tab:main}, fine-tuned or not.
 
\paragraph{The decomposed term closes the gap to holistic scoring.}
Switching the scoring rule from \emph{Faith} (Soft-TIFA-AM) to \emph{Agree} narrows the gap to VQAScore on GenAI-Bench from $0.15$ to just $0.03$ SRCC (Table~\ref{tab:anchor}). This shows that expected agreement alone drives the improvement, while also raising NegBench accuracy from below chance to $88.3\%$ without hurting performance on TIFA160.
 
\paragraph{Merging takes the best of both.}
On all three correlation benchmarks \textsc{\model} matches whichever term suits the benchmark and improves on it (Table~\ref{tab:anchor}). The gain is significant on GenAI-Bench and positive but within the interval on TIFA160 and RichHF. The improvement comes from the low end: the merged metric's lowest-scored items are rated worse by humans than VQAScore's on every benchmark and backbone, while the top tails coincide (Appendix~\ref{ann:calib}).

\paragraph{The merge carries no penalty where one term suffices.} On Winoground's short compositional captions, the decomposed term adds little beyond what the holistic pass captures, keeping~\textsc{\model} within the confidence interval of VQAScore across all three accuracies (Table~\ref{tab:anchor}). Furthermore, soft logit scoring across both terms (\emph{Agree}) drives performance far more than decomposition itself, leaving discrete one-hot approaches like DSG lagging by 20 points.

\begin{table*}[t]
\centering
\small
\setlength{\tabcolsep}{1pt}
\renewcommand{\arraystretch}{1}
\resizebox{\linewidth}{!}{%
\begin{tabular}{l ccc ccc c}
\toprule
\multirow{2}{*}{\textbf{Method}} & \multicolumn{3}{c}{\textbf{Human corr. (SRCC)}} & \multicolumn{3}{c}{\textbf{Winoground (acc.\ \%)}} & \textbf{NegBench} \\
\cmidrule(lr){2-4}\cmidrule(lr){5-7}\cmidrule(lr){8-8}
 & GenAI-B & TIFA160 & RichHF & Text & Image & Group & acc.\ \% \\
\midrule
DSG            & 0.549 & 0.673 & 0.594 & 52.0 & 46.5 & 44.5 & 34.7 \\
Soft-TIFA-AM (\emph{Faith})  & 0.546 & 0.762 & 0.636 & 75.8 & 71.8 & 64.8 & 19.1 \\
VQAScore (\emph{Holistic})       & \underline{0.698} & 0.737 & \underline{0.656} & \textbf{83.2} & \textbf{82.8} & \textbf{75.5} & 84.7 \\
\midrule
\emph{Agree} & 0.671 & \underline{0.763} & 0.612 & 74.0 & 74.0 & 65.8 & \textbf{88.3} \\
\textsc{\model}                  & \textbf{0.720} & \textbf{0.780} & \textbf{0.667} & \underline{82.2} & \textbf{82.8} & \underline{74.5} & \underline{86.9} \\
\midrule
$\Delta$ vs.\ best baseline & $+$.022 & $+$.018 & $+$.011 & $-$1.0 & $+$0.0 & $-$1.0 & $+$2.3 \\
\quad {\scriptsize [95\% CI]}
  & {\scriptsize [$+$.016,\,$+$.028]} 
  & {\scriptsize [$-$.002,\,$+$.040]} 
  & {\scriptsize [$-$.013,\,$+$.031]}
  & {\scriptsize [$-$3.5,\,$+$1.0]} 
  & {\scriptsize [$-$3.0,\,$+$2.5]} 
  & {\scriptsize [$-$3.8,\,$+$1.3]}
  & {\scriptsize [$+$1.8,\,$+$2.8]} \\
\bottomrule
\end{tabular}
}
\caption{\textbf{Fixed-backbone comparison on Qwen3.5-27B.} All methods share images, prompts, and decoding. \emph{Agree} differs from \emph{Faith} only in scoring rule; \textsc{\model} combines it with holistic. $\Delta$ gives the difference of \textsc{\model} over the strongest baseline in each column, with its paired BCa 95\% CI.}
\label{tab:anchor}
\end{table*}

\begin{figure}[h]
    \centering
    \includegraphics[width=\textwidth]{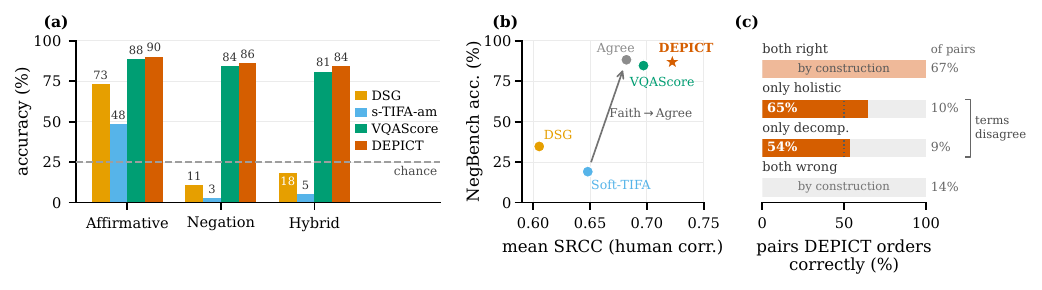}
         \caption{\textbf{Why agreement and merging work (Qwen3.5-27B).} \textbf{(a)} NegBench accuracy by template; VQAScore is unaffected because its single question contains the negation. \textbf{(b)} NegBench accuracy vs.\ mean SRCC; \emph{Agree} improves on \emph{Faith}, and \textsc{\model} boosts correlation further. \textbf{(c)} GenAI-Bench item pairs grouped by term alignment.}
    \label{fig:analysis}
\end{figure}

\subsection{Why do agreement and merging work?}
\label{sec:analysis}

In this section, we analyze how \emph{Agreement} overcomes the structural flaws of fixed-\textsc{yes} rules on negation, while merging recovers performance on contested samples where terms disagree.

\paragraph{The fixed-\textsc{yes} assumption fails on negation.}
By reading the expected answer from the caption, \textsc{\model} is the only decomposed metric accurate across all three templates (Fig.~\ref{fig:analysis}a), while DSG and Soft-TIFA lose $85$--$95\%$ of their accuracy on negated items and fall below chance. This occurs because $87\%$ of questions for negated captions and $47\%$ for hybrid ones expect \textsc{no} vs $4\%$ elsewhere (Appendix~\ref{ann:yesno}), so a fixed-\textsc{yes} rule gives a low score to every correct \textsc{no} before looking at the image, penalizing faithful images (Appendix~\ref{ann:negbench}-\ref{ann:skills};  text reference reliability in Appendix~\ref{ann:reference}).

\paragraph{The terms are complementary.}
\emph{Agreement} alone lifts both NegBench accuracy and human correlation over \emph{Faith} but trails VQAScore on correlation; merging both terms places \textsc{\model} ahead on both (Fig.~\ref{fig:analysis}b). The merge of both terms (Eq.~\ref{eq:merge}) works because the terms fail on different samples: on GenAI-Bench they disagree on $19\%$ of item pairs. Fig.~\ref{fig:analysis}c shows that \textsc{\model} tracks the results where both terms agree and it follows whichever term separates the two images more strongly where they disagree. \textsc{\model} keeps $65\%$ of the pairs only the holistic term orders correctly and $54\%$ of those only the decomposed term does. Whereas individual terms fail on each other's exclusive wins, \textsc{\model} achieves $60\%$ accuracy across contested pairs, easily outperforming the holistic ($53\%$) and decomposed ($47\%$) terms alone. The merge relies on agreement: combining VQAScore with Soft-TIFA instead leaves GenAI-Bench unchanged and lowers NegBench to $62.2$, below VQAScore alone, since the fixed-\textsc{yes} term drags down the holistic score on negated items (Appendix~\ref{ann:merge-ablation}).

\begin{figure}[h]
\centering
\begin{subfigure}{0.9\linewidth}
    \centering
    \includegraphics[width=\linewidth]{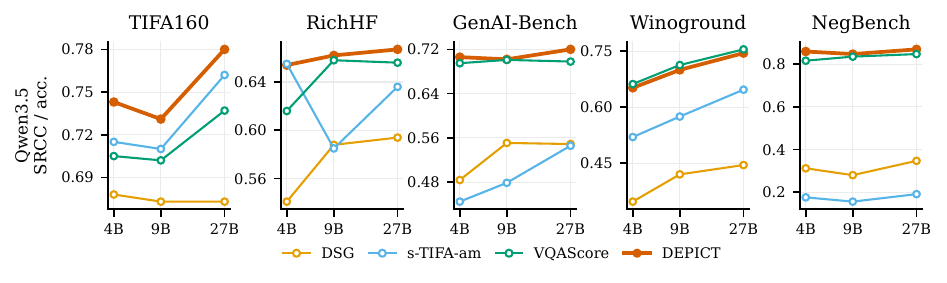}
    \caption{Absolute scores along the Qwen3.5 size ladder.}
    \label{fig:scaling}
\end{subfigure}
\begin{subfigure}{0.9\linewidth}
    \centering
    \includegraphics[width=\linewidth]{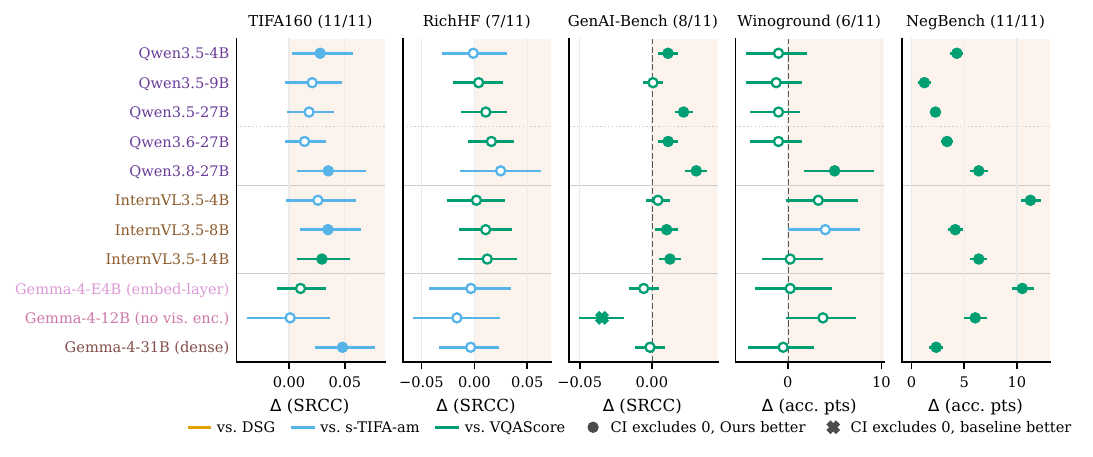}
    \caption{Difference to the strongest baseline on eleven backbones.}
    \label{fig:forest}
\end{subfigure}
\caption{\textbf{Training-free metrics across backbones.} Evaluated on five benchmarks (SRCC for correlation benchmarks; accuracy on Winoground and NegBench). (a) Other families in Appendix~\ref{ann:sweep}. (b) Score differences ($\Delta$) between \textsc{\model} and the strongest baseline per backbone (paired bootstrap 95\% CI). Values $>0$ favor \textsc{\model}; solid markers exclude zero, and crosses indicate significant losses. Colors denote the baseline and titles wins count.}
\label{fig:backbones}
\end{figure}

\subsection{How do metrics behave across backbones?}
\label{sec:scaling}

The backbone of a training-free metric is a free choice, so we ask whether \textsc{\model}'s advantage depends on it. Using the eleven-backbone sweep described in Sec.~\ref{sec:backbones}, Fig.~\ref{fig:scaling} traces Qwen3.5 size scaling, while Fig.~\ref{fig:forest} reports baseline differences across family, architecture, and generation axes.

\paragraph{Size helps where the backbone is the bottleneck.}
Along the Qwen3.5 ladder, \textsc{\model} leads or ties at every size on all three correlation benchmarks, where larger backbones mainly narrow the gap for the weaker decomposed baselines. On NegBench its advantage does not depend on size, as accuracy holds at $85$--$87\%$ at every size while fixed-\textsc{yes} baselines stay at or below chance, showing that how the score is computed matters more than backbone size. In contrast, on Winoground all methods gain similarly, suggesting that the backbone is the bottleneck.

\paragraph{The advantage is consistent across backbones.}
Across all eleven backbones (Fig.~\ref{fig:forest}), \textsc{\model} matches or beats the strongest baseline in 54 out of 55 backbone--benchmark pairs, winning in 43 cases with 23 being statistically significant. Across families, its NegBench gains are larger on InternVL3.5 than on Qwen3.5 ($4$--$11$ points against $1$--$4$), even though InternVL3.5 is far weaker on Winoground at matched size, so agreement scoring compensates for a weaker backbone. Across architectures, \textsc{\model}'s only loss occurs on GenAI-Bench with the Gemma-4-12B variant without a vision encoder, the weakest backbone on every benchmark (Appendix~\ref{ann:sweep}). Across generations, its margins are largest on the newest backbone, Qwen3.8-27B, significant on four out of five benchmarks, even though that backbone scores below its predecessors for nearly every method.

\subsection{Which components matter?}
\label{sec:ablations}
Table~\ref{tab:ablations} ablates the design choices of the decomposed term on Qwen3.5-27B, replacing one component at a time while holding the rest at the final configuration and testing multiple scoring-rules. We ablate reasoning in the answering step in Appendix~\ref{ann:reasoning}.

\begin{table}[t]
\centering
\small
\setlength{\tabcolsep}{3pt}
\begin{subtable}[t]{0.56\linewidth}
\centering
\begin{tabular}{llccc}
\toprule
\textbf{Comp.} & \textbf{Setting} & \textbf{Wino.} & \textbf{GenAI} & \textbf{Neg.} \\
\midrule
Questions & via tuples & 44.8 & 0.573 & 64.4 \\
\multirow{3}{*}{Filter} & no $w_k$            & 65.0 & 0.668 & \textbf{88.3} \\
                        & pruning, no $w_k$   & 62.7 & 0.642 & 88.0 \\
                        & pruning + $w_k$     & 63.2 & 0.644 & 88.0 \\
\midrule
\multicolumn{2}{l}{\textbf{Default}} & \textbf{65.8} & \textbf{0.671} & \textbf{88.3} \\
\bottomrule
\end{tabular}
\caption{Pipeline components.}
\label{tab:abl-pipeline}
\end{subtable}\hfill
\begin{subtable}[t]{0.40\linewidth}
\centering
\begin{tabular}{lccc}
\toprule
\textbf{Rule} & \textbf{Wino.} & \textbf{GenAI} & \textbf{Neg.} \\
\midrule
Faith, hard$^{\ddagger}$ & 48.5 & 0.513 & 21.9 \\
Faith$^{\dagger}$        & 64.8 & 0.546 & 19.1 \\
Confirm                  & 64.0 & 0.624 & 66.4 \\
\textit{Agree}           & 65.0 & 0.668 & 88.3 \\
\midrule
\quad + $w_k$ (default)  & \textbf{65.8} & \textbf{0.671} & \textbf{88.3} \\
\bottomrule
\end{tabular}
\caption{Scoring rule on identical questions.}
\label{tab:abl-scoring}
\end{subtable}
\caption{\textbf{Component ablations of the decomposed term (Qwen3.5-27B).} Wino.: Winoground group accuracy; GenAI: GenAI-Bench SRCC; Neg.: NegBench accuracy; higher is better. (a) Default: questions generated directly from the caption, \emph{Agree} scoring, commitment weighting $w_k$~\eqref{eq:cw}; each row changes one component. (b) Rules without $w_k$, then the default. $^\dagger$Identical to Soft-TIFA-AM. $^\ddagger$DSG's hard fixed-\textsc{yes} rule on our questions, without its dependency graph.}
\label{tab:ablations}
\vspace{-1.5em}
\end{table}

\paragraph{Tuple Extraction.}
Replicating DSG's tuple extraction hurts all benchmarks (Table~\ref{tab:ablations}a, top), 47\% of negated NegBench captions yield no tuples and Negation-subset accuracy falls to $4.1$, six times below chance, showing it cannot express absence. On Winoground pairs, which share entities but differ in structure, it generates overlapping questions (Jaccard $0.44$ vs.\ $0.27$ without tuples; 13\% identical), shrinking the paired score margin by ${\sim}40\%$ and costing 21 group-accuracy points (Appendix~\ref{ann:tuples}). Generating questions directly from the caption preserves both.

\paragraph{Question pruning.}
\label{sec:abl-pruning}
Post-hoc pruning lowers performance on every benchmark. Earlier pipelines filtered questions to remove noise~\citep{Cho_2024_davidsonian}. Agreement scoring handles this per question: a question the two channels cannot settle scores near the midpoint rather than being rewarded or silenced, and $w_k$ down-weights it (Appendix~\ref{ann:weighting}), so pruning only removes questions carrying signal.

\paragraph{Scoring.} The scoring rule is the largest single factor (Table~\ref{tab:ablations}b), discretizing Faith as DSG does costs 16.3 Winoground points and 0.033 SRCC, while Faith stays below chance on NegBench. Agree raises it $4.6\times$ and GenAI-Bench SRCC by 23\%, moving Winoground by at most one point.

\section{Conclusions}
\label{sec:conclusions}

Fine-tuned, holistic, and decomposed metrics each capture distinct part of caption-image alignment: fine-tuning aligns with human judgments but ties the evaluator to one backbone; holistic VQA preserves global context but misses fine-grained errors; and standard decomposition uncovers compositional logic but relies on a fixed-\textsc{yes} target that breaks on negated captions. \textsc{\model} resolves these trade-offs without training by evaluating two-sided answer agreement weighted by caption commitment, merged with a holistic VQA pass. Agreement lifts NegBench accuracy from $19\%$ to $88\%$, and the merged score matches or exceeds top evaluators across five benchmarks and eleven backbones. On DynEval's matched Qwen3-VL-4B backbone, \textsc{\model} beats DynEval-4B on GenAI-Bench ($0.630$ vs.\ $0.595$ SRCC) and RichHF ($0.607$ vs.\ $0.546$), though fine-tuning retains a lead on TIFA160 ($0.802$ vs.\ $0.691$).

Limitations remain: decomposition brings no gain on short prompts that hinge on compositional word order, such as Winoground, although enabling  reasoning on answers improves results there; and because both channels share one backbone, a prior shared by both can make the same hallucination look like agreement. Answering each channel with a different backbone would decouple these priors, at the cost of hosting two models.

{\small
\bibliographystyle{plainnat}
\bibliography{main}

@String(CVPR= {IEEE Conf. Comput. Vis. Pattern Recog.})

@String(ICCV= {Int. Conf. Comput. Vis.})

@String(ECCV= {Eur. Conf. Comput. Vis.})

@String(ICLR = {Int. Conf. Learn. Represent.})

@String(AAAI = {AAAI})

@String(CVPR  = {CVPR})

@String(ICCV  = {ICCV})

@String(ECCV  = {ECCV})

@String(ICLR  = {ICLR})

@inproceedings{Hu_2023_tifa,
  author       = {Yushi Hu and
                  Benlin Liu and
                  Jungo Kasai and
                  Yizhong Wang and
                  Mari Ostendorf and
                  Ranjay Krishna and
                  Noah A. Smith},
  title        = {{TIFA:} Accurate and Interpretable Text-to-Image Faithfulness Evaluation
                  with Question Answering},
  booktitle    = {{IEEE/CVF} International Conference on Computer Vision, {ICCV} 2023,
                  Paris, France, October 1-6, 2023},
  pages        = {20349--20360},
  publisher    = {{IEEE}},
  year         = {2023},
  url          = {https://doi.org/10.1109/ICCV51070.2023.01866},
  doi          = {10.1109/ICCV51070.2023.01866},
  bibsource    = {dblp computer science bibliography, https://dblp.org}
}

@inproceedings{vq2,
  author       = {Michal Yarom and
                  Yonatan Bitton and
                  Soravit Changpinyo and
                  Roee Aharoni and
                  Jonathan Herzig and
                  Oran Lang and
                  Eran Ofek and
                  Idan Szpektor},
  editor       = {Alice Oh and
                  Tristan Naumann and
                  Amir Globerson and
                  Kate Saenko and
                  Moritz Hardt and
                  Sergey Levine},
  title        = {What You See is What You Read? Improving Text-Image Alignment Evaluation},
  booktitle    = {Advances in Neural Information Processing Systems 36: Annual Conference
                  on Neural Information Processing Systems 2023, NeurIPS 2023, New Orleans,
                  LA, USA, December 10 - 16, 2023},
  year         = {2023},
  url          = {http://papers.nips.cc/paper\_files/paper/2023/hash/056e8e9c8ca9929cb6cf198952bf1dbb-Abstract-Conference.html},
  bibsource    = {dblp computer science bibliography, https://dblp.org}
}

@inproceedings{vqascore,
  author       = {Zhiqiu Lin and
                  Deepak Pathak and
                  Baiqi Li and
                  Jiayao Li and
                  Xide Xia and
                  Graham Neubig and
                  Pengchuan Zhang and
                  Deva Ramanan},
  editor       = {Ales Leonardis and
                  Elisa Ricci and
                  Stefan Roth and
                  Olga Russakovsky and
                  Torsten Sattler and
                  G{\"{u}}l Varol},
  title        = {Evaluating Text-to-Visual Generation with Image-to-Text Generation},
  booktitle    = {Computer Vision - {ECCV} 2024 - 18th European Conference, Milan, Italy,
                  September 29-October 4, 2024, Proceedings, Part {IX}},
  series       = {Lecture Notes in Computer Science},
  volume       = {15067},
  pages        = {366--384},
  publisher    = {Springer},
  year         = {2024},
  url          = {https://doi.org/10.1007/978-3-031-72673-6\_20},
  doi          = {10.1007/978-3-031-72673-6\_20},
  bibsource    = {dblp computer science bibliography, https://dblp.org}
}

@inproceedings{Cho_2024_davidsonian,
  author       = {Jaemin Cho and
                  Yushi Hu and
                  Jason M. Baldridge and
                  Roopal Garg and
                  Peter Anderson and
                  Ranjay Krishna and
                  Mohit Bansal and
                  Jordi Pont{-}Tuset and
                  Su Wang},
  title        = {Davidsonian Scene Graph: Improving Reliability in Fine-grained Evaluation
                  for Text-to-Image Generation},
  booktitle    = {The Twelfth International Conference on Learning Representations,
                  {ICLR} 2024, Vienna, Austria, May 7-11, 2024},
  publisher    = {OpenReview.net},
  year         = {2024},
  url          = {https://openreview.net/forum?id=ITq4ZRUT4a},
  bibsource    = {dblp computer science bibliography, https://dblp.org}
}

@misc{Han_2024_evalmuse40k,
  title = {{{EvalMuse-40K}}: {{A Reliable}} and {{Fine-Grained Benchmark}} with {{Comprehensive Human Annotations}} for {{Text-to-Image Generation Model Evaluation}}},
  author = {Han, Shuhao and Fan, Haotian and Fu, Jiachen and Li, Liang and Li, Tao and Cui, Junhui and Wang, Yunqiu and Tai, Yang and Sun, Jingwei and Guo, Chunle and Li, Chongyi},
  year = 2024,
  number = {arXiv:2412.18150},
  eprint = {2412.18150},
  primaryclass = {cs.CV},
  doi = {10.48550/arXiv.2412.18150},
  archiveprefix = {arXiv}
}

@inproceedings{Hessel_2021_clipscore,
  title = {{{CLIPScore}}: {{A Reference-free Evaluation Metric}} for {{Image Captioning}}},
  booktitle = {Conference on {{Empirical Methods}} in {{Natural Language Processing}}},
  author = {Hessel, Jack and Holtzman, Ari and Forbes, Maxwell and Bras, Ronan Le and Choi, Yejin},
  year = 2021,
  eprint = {2104.08718},
  primaryclass = {cs.CV},
  doi = {10.18653/v1/2021.emnlp-main.595},
  archiveprefix = {arXiv}
}

@misc{Kamath_2025_geneval2,
  title = {{{GenEval}} 2: {{Addressing Benchmark Drift}} in {{Text-to-Image Evaluation}}},
  author = {Kamath, Amita and Chang, Kai-Wei and Krishna, Ranjay and Zettlemoyer, Luke and Hu, Yushi and Ghazvininejad, Marjan},
  year = 2025,
  number = {arXiv:2512.16853},
  eprint = {2512.16853},
  primaryclass = {cs.CV},
  doi = {10.48550/arXiv.2512.16853},
  archiveprefix = {arXiv}
}

@inproceedings{Kirstain_2023_pickapic,
  title = {Pick-a-{{Pic}}: {{An Open Dataset}} of {{User Preferences}} for {{Text-to-Image Generation}}},
  booktitle = {Neural {{Information Processing Systems}}},
  author = {Kirstain, Yuval and Polyak, Adam and Singer, Uriel and Matiana, Shahbuland and Penna, Joe and Levy, Omer},
  year = 2023,
  eprint = {2305.01569},
  primaryclass = {cs.CV},
  doi = {10.48550/arXiv.2305.01569},
  archiveprefix = {arXiv}
}

@InProceedings{Marjit_2026_dyneval,
author="Marjit, Shyam
and Baiju, Dheeraj
and Shikarkhane, Anuj
and Sakthieswaran, Akhil
and Paul, Sayak
and Chakraborty, Anirban",
editor="Favaro, Paolo
and Kukelova, Zuzana
and Maki, Atsuto
and Rohrbach, Anna
and Schindler, Konrad
and Tombari, Federico",
title="DynEval: Holistic Evaluations of T2I Generative Models in the Wild",
booktitle="Computer Vision -- ECCV 2026",
year="2026",
publisher="Springer Nature Switzerland",
address="Cham",
pages="417--435",
isbn="978-3-032-37010-5"
}

@misc{Wang_2026_unigenbench,
  title = {{{UniGenBench}}++: {{A Unified Semantic Evaluation Benchmark}} for {{Text-to-Image Generation}}},
  author = {Wang, Yibin and Li, Zhimin and Zang, Yuhang and Bu, Jiazi and Zhou, Yujie and Xin, Yi and He, Junjun and Wang, Chunyu and Lu, Qinglin and Jin, Cheng and Wang, Jiaqi},
  year = 2026,
  number = {arXiv:2510.18701},
  eprint = {2510.18701},
  primaryclass = {cs.CV},
  doi = {10.48550/arXiv.2510.18701},
  archiveprefix = {arXiv}
}

@misc{Wu_2023_human,
  title = {Human {{Preference Score}} v2: {{A Solid Benchmark}} for {{Evaluating Human Preferences}} of {{Text-to-Image Synthesis}}},
  author = {Wu, Xiaoshi and Hao, Yiming and Sun, Keqiang and Chen, Yixiong and Zhu, Feng and Zhao, Rui and Li, Hongsheng},
  year = 2023,
  number = {arXiv:2306.09341},
  eprint = {2306.09341},
  primaryclass = {cs.CV},
  doi = {10.48550/arXiv.2306.09341},
  archiveprefix = {arXiv}
}

@inproceedings{Xu_2023_imagereward,
  title = {{{ImageReward}}: {{Learning}} and {{Evaluating Human Preferences}} for {{Text-to-Image Generation}}},
  booktitle = {Neural {{Information Processing Systems}}},
  author = {Xu, Jiazheng and Liu, Xiao and Wu, Yuchen and Tong, Yuxuan and Li, Qinkai and Ding, Ming and Tang, Jie and Dong, Yuxiao},
  year = 2023,
  eprint = {2304.05977},
  primaryclass = {cs.CV},
  doi = {10.48550/arXiv.2304.05977},
  archiveprefix = {arXiv}
}

@inproceedings{Yang_2025_longt2ibench,
  author       = {Zhichao Yang and
                  Tianjiao Gu and
                  Jianjie Wang and
                  Feiyu Lin and
                  Xiangfei Sheng and
                  Pengfei Chen and
                  Leida Li},
  title        = {LongT2IBench: {A} Benchmark for Evaluating Long Text-to-Image Generation with Graph-structured Annotations},
  booktitle    = {{AAAI}},
  pages        = {11820--11828},
  publisher    = {{AAAI} Press},
  year         = {2026}
}

@inproceedings{tu-etal-2025-automatic,
    title = "Automatic Evaluation for Text-to-image Generation: Task-decomposed Framework, Distilled Training, and Meta-evaluation Benchmark",
    author = "Tu, Rong-Cheng  and
      Ma, Zi-Ao  and
      Lan, Tian  and
      Zhao, Yuehao  and
      Huang, Heyan  and
      Mao, Xian-Ling",
    editor = "Che, Wanxiang  and
      Nabende, Joyce  and
      Shutova, Ekaterina  and
      Pilehvar, Mohammad Taher",
    booktitle = "Proceedings of the 63rd Annual Meeting of the Association for Computational Linguistics (Volume 1: Long Papers)",
    month = jul,
    year = "2025",
    address = "Vienna, Austria",
    publisher = "Association for Computational Linguistics",
    url = "https://aclanthology.org/2025.acl-long.1088/",
    doi = "10.18653/v1/2025.acl-long.1088",
    pages = "22340--22361",
    ISBN = "979-8-89176-251-0"
}

@article{flux,
  title={Flux. 1 kontext: Flow matching for in-context image generation and editing in latent space},
  author={Labs, Black Forest and Batifol, Stephen and Blattmann, Andreas and Boesel, Frederic and Consul, Saksham and Diagne, Cyril and Dockhorn, Tim and English, Jack and English, Zion and Esser, Patrick and others},
  journal={arXiv preprint arXiv:2506.15742},
  year={2025}
}

@article{chefer2026self,
  title={Self-supervised flow matching for scalable multi-modal synthesis},
  author={Chefer, Hila and Esser, Patrick and Lorenz, Dominik and Podell, Dustin and Raja, Vikash and Tong, Vinh and Torralba, Antonio and Rombach, Robin},
  journal={arXiv preprint arXiv:2603.06507},
  year={2026}
}

@article{qwenimage,
  title={Qwen-image-2.0 technical report},
  author={Zhao, Bing and Wu, Chenfei and Li, Deqing and Meng, Hao and Li, Jiahao and Zhang, Jie and Zhou, Jingren and Lin, Junyang and Gao, Kaiyuan and Cao, Kuan and others},
  journal={arXiv preprint arXiv:2605.10730},
  year={2026}
}

@article{yang2025qwen3,
  title={Qwen3 technical report},
  author={Yang, An and Li, Anfeng and Yang, Baosong and Zhang, Beichen and Hui, Binyuan and Zheng, Bo and Yu, Bowen and Gao, Chang and Huang, Chengen and Lv, Chenxu and others},
  journal={arXiv preprint arXiv:2505.09388},
  year={2025}
}

@article{wang2025internvl3,
  title={Internvl3. 5: Advancing open-source multimodal models in versatility, reasoning, and efficiency},
  author={Wang, Weiyun and Gao, Zhangwei and Gu, Lixin and Pu, Hengjun and Cui, Long and Wei, Xingguang and Liu, Zhaoyang and Jing, Linglin and Ye, Shenglong and Shao, Jie and others},
  journal={arXiv preprint arXiv:2508.18265},
  year={2025}
}

@article{team2026gemma,
  title={Gemma 4 technical report},
  author={Team, Gemma and Abd, Sherif El and Aggarwal, Vaibhav and Algayres, Robin and Andreev, Alek and Bachem, Olivier and Ballantyne, Ian and Brick, Cormac and C{\u{a}}rbune, Victor and Casbon, Michelle and others},
  journal={arXiv preprint arXiv:2607.02770},
  year={2026}
}

@inproceedings{
baiqi-li2024genaibench,
title={Gen{AI}-Bench: A Holistic Benchmark for Compositional Text-to-Visual Generation},
author={Baiqi Li and Zhiqiu Lin and Deepak Pathak and Jiayao Emily Li and Xide Xia and Graham Neubig and Pengchuan Zhang and Deva Ramanan},
booktitle={Synthetic Data for Computer Vision Workshop @ CVPR 2024},
year={2024},
url={https://openreview.net/forum?id=hJm7qnW3ym}
}

@inproceedings{richhf,
  title={Rich Human Feedback for Text-to-Image Generation},
  author={Youwei Liang and Junfeng He and Gang Li and Peizhao Li and Arseniy Klimovskiy and Nicholas Carolan and Jiao Sun and Jordi Pont-Tuset and Sarah Young and Feng Yang and Junjie Ke and Krishnamurthy Dj Dvijotham and Katie Collins and Yiwen Luo and Yang Li and Kai J Kohlhoff and Deepak Ramachandran and Vidhya Navalpakkam},
  booktitle={Proceedings of the IEEE/CVF Conference on Computer Vision and Pattern Recognition},
  year={2024},
}

@inproceedings{kogilathota2026halp,
  title={HALP: Detecting hallucinations in vision-language models without generating a single token},
  author={Kogilathota, Sai Akhil and EG, Sripadha Vallabha and Sun, Luzhe and Zhou, Jiawei},
  booktitle={Proceedings of the 19th Conference of the European Chapter of the Association for Computational Linguistics (Volume 1: Long Papers)},
  pages={6067--6085},
  year={2026}
}

@inproceedings{noisyclip,
  author       = {Vasco Ramos and
                  Regev Cohen and
                  Idan Szpektor and
                  Jo{\~{a}}o Magalh{\~{a}}es},
  editor       = {Paolo Favaro and
                  Zuzana Kukelova and
                  Atsuto Maki and
                  Anna Rohrbach and
                  Konrad Schindler and
                  Federico Tombari},
  title        = {Early Estimation of Language to Latent Alignment in Diffusion Models},
  booktitle    = {Computer Vision - {ECCV} 2026 - 19th European Conference, Malm{\"{o}},
                  Sweden, September 8-12, 2026, Proceedings, Part {XX}},
  series       = {Lecture Notes in Computer Science},
  volume       = {17020},
  pages        = {625--643},
  publisher    = {Springer},
  year         = {2026},
  url          = {https://doi.org/10.1007/978-3-032-37553-7\_36},
  doi          = {10.1007/978-3-032-37553-7\_36},
  bibsource    = {dblp computer science bibliography, https://dblp.org}
}

@article{schuhmann2021laion,
  title={Laion-400m: Open dataset of clip-filtered 400 million image-text pairs},
  author={Schuhmann, Christoph and Vencu, Richard and Beaumont, Romain and Kaczmarczyk, Robert and Mullis, Clayton and Katta, Aarush and Coombes, Theo and Jitsev, Jenia and Komatsuzaki, Aran},
  journal={arXiv preprint arXiv:2111.02114},
  year={2021}
}

@inproceedings{
gadre2023datacomp,
title={DataComp: In search of the next generation of multimodal datasets},
author={Samir Yitzhak Gadre and Gabriel Ilharco and Alex Fang and Jonathan Hayase and Georgios Smyrnis and Thao Nguyen and Ryan Marten and Mitchell Wortsman and Dhruba Ghosh and Jieyu Zhang and Eyal Orgad and Rahim Entezari and Giannis Daras and Sarah M Pratt and Vivek Ramanujan and Yonatan Bitton and Kalyani Marathe and Stephen Mussmann and Richard Vencu and Mehdi Cherti and Ranjay Krishna and Pang Wei Koh and Olga Saukh and Alexander Ratner and Shuran Song and Hannaneh Hajishirzi and Ali Farhadi and Romain Beaumont and Sewoong Oh and Alex Dimakis and Jenia Jitsev and Yair Carmon and Vaishaal Shankar and Ludwig Schmidt},
booktitle={Thirty-seventh Conference on Neural Information Processing Systems Datasets and Benchmarks Track},
year={2023},
url={https://openreview.net/forum?id=dVaWCDMBof}
}

@inproceedings{black2024trainingddpo,
  title={Training diffusion models with reinforcement learning},
  author={Black, Kevin and Janner, Michael and Du, Yilun and Kostrikov, Ilya and Levine, Sergey},
  booktitle={International Conference on Learning Representations},
  volume={2024},
  pages={4965--4987},
  year={2024}
}

@inproceedings{thrush2022winoground,
  title={Winoground: Probing vision and language models for visio-linguistic compositionality},
  author={Thrush, Tristan and Jiang, Ryan and Bartolo, Max and Singh, Amanpreet and Williams, Adina and Kiela, Douwe and Ross, Candace},
  booktitle={2022 IEEE/CVF Conference on Computer Vision and Pattern Recognition (CVPR)},
  pages={5228--5238},
  year={2022},
  organization={IEEE}
}

@inproceedings{alhamoud2025vision_negbench,
  title={Vision-language models do not understand negation},
  author={Alhamoud, Kumail and Alshammari, Shaden and Tian, Yonglong and Li, Guohao and Torr, Philip HS and Kim, Yoon and Ghassemi, Marzyeh},
  booktitle={2025 IEEE/CVF Conference on Computer Vision and Pattern Recognition (CVPR)},
  pages={29612--29622},
  year={2025},
  organization={IEEE}
}

@inproceedings{li2023blip2,
  title={Blip-2: Bootstrapping language-image pre-training with frozen image encoders and large language models},
  author={Li, Junnan and Li, Dongxu and Savarese, Silvio and Hoi, Steven},
  booktitle={International conference on machine learning},
  pages={19730--19742},
  year={2023},
  organization={PmLR}
}

@inproceedings{Yuksekgonul0KJ023_bag_of_words,
  author       = {Mert Y{\"{u}}ksekg{\"{o}}n{\"{u}}l and
                  Federico Bianchi and
                  Pratyusha Kalluri and
                  Dan Jurafsky and
                  James Zou},
  title        = {When and Why Vision-Language Models Behave like Bags-Of-Words, and
                  What to Do About It?},
  booktitle    = {The Eleventh International Conference on Learning Representations,
                  {ICLR} 2023, Kigali, Rwanda, May 1-5, 2023},
  publisher    = {OpenReview.net},
  year         = {2023},
  url          = {https://openreview.net/forum?id=KRLUvxh8uaX},
  bibsource    = {dblp computer science bibliography, https://dblp.org}
}

@inproceedings{
koishigarina2026clip_bag_of_words,
title={{CLIP} Behaves like a Bag-of-Words Model Cross-modally but not Uni-modally},
author={Darina Koishigarina and Arnas Uselis and Seong Joon Oh},
booktitle={The Fourteenth International Conference on Learning Representations},
year={2026},
url={https://openreview.net/forum?id=DldwXCCP25}
}

@inproceedings{wiles2024gecko,
  author       = {Olivia Wiles and
                  Chuhan Zhang and
                  Isabela Albuquerque and
                  Ivana Kajic and
                  Su Wang and
                  Emanuele Bugliarello and
                  Yasumasa Onoe and
                  Pinelopi Papalampidi and
                  Ira Ktena and
                  Christopher Knutsen and
                  Cyrus Rashtchian and
                  Anant Nawalgaria and
                  Jordi Pont{-}Tuset and
                  Aida Nematzadeh},
  title        = {Revisiting text-to-image evaluation with Gecko: on metrics, prompts,
                  and human rating},
  booktitle    = {{ICLR}},
  publisher    = {OpenReview.net},
  year         = {2025}
}

@inproceedings{ku2024imagenhub,
  title={Imagenhub: Standardizing the evaluation of conditional image generation models},
  author={Ku, Max and Li, Tianle and Zhang, Kai and Lu, Yujie and Fu, Xingyu and Zhuang, Wenwen and Chen, Wenhu},
  booktitle={International Conference on Learning Representations},
  volume={2024},
  pages={46689--46722},
  year={2024}
}

@article{singh2023divide,
  title={Divide, evaluate, and refine: Evaluating and improving text-to-image alignment with iterative vqa feedback},
  author={Singh, Jaskirat and Zheng, Liang},
  journal={Advances in Neural Information Processing Systems},
  volume={36},
  pages={70799--70811},
  year={2023}
}
}
\newpage
\appendix
\crefalias{section}{appendix}
\crefalias{subsection}{appendix}
\crefalias{subsubsection}{appendix}

\setcounter{secnumdepth}{-2}
\part{Appendix}
\setcounter{secnumdepth}{3}
\etocsetnexttocdepth{subsection}
\localtableofcontents
\clearpage

\appendix
\section{Experimental details}
\label{ann:details}

\subsection{Benchmarks}
\label{ann:datasets}

Table~\ref{tab:datasets} summarizes the five benchmarks. GenAI-Bench~\citep{baiqi-li2024genaibench} pairs compositional prompts with images from several generators, each rated for alignment on a 1-5 scale. TIFA160~\citep{Hu_2023_tifa} contains $160$ prompts with five generated images each, rated on the same scale. For RichHF-18K~\citep{richhf} we use the test split and its per-image misalignment score. On all three we report SRCC on raw scores and PLCC after a four-parameter logistic fit, pooled over image-caption pairs.

\begin{table}[!htbp]
\centering
\small
\setlength{\tabcolsep}{4pt}
\resizebox{\linewidth}{!}{%
\begin{tabular}{l p{3.6cm} l l l c}
\toprule
\textbf{Benchmark} & \textbf{Source} & \textbf{Evaluated units} & \textbf{Human signal} & \textbf{Metric} & $\bar N$ \textbf{(max)} \\
\midrule
\multicolumn{6}{l}{\textit{Correlation with human ratings}} \\
GenAI-Bench & GenAI-Bench prompts and generated images & 9{,}600 pairs, 1{,}600 prompts & 1--5 alignment & SRCC, PLCC & 5.47 (13) \\
TIFA160     & TIFA160 prompts, 5 images each & 800 pairs, 160 prompts & 1--5 alignment & SRCC, PLCC & 4.89 (11) \\
RichHF      & RichHF-18K test split & 955 pairs, 493 captions & misalignment score & SRCC, PLCC & 5.52 (32) \\
\midrule
\multicolumn{6}{l}{\textit{Accuracy}} \\
Winoground  & Winoground & 400 examples, 800 matched pairs & correct pairing & text / image / group acc. & 4.16 (10) \\
NegBench    & COCO-MCQ split (COCO val2017) & 5{,}914 images $\times$ 4 captions & correct caption & 4-way MCQ acc. & 1.92 (6) \\
\bottomrule
\end{tabular}}
\caption{\textbf{Benchmarks.} PLCC is computed after a four-parameter logistic fit. NegBench captions follow three templates (affirmative, negation, hybrid); each image has one correct caption and three distractors. $\bar N$: mean number of generated questions per unique caption, with the maximum in parentheses.}
\label{tab:datasets}
\end{table}

Winoground~\citep{thrush2022winoground} contains $400$ examples of two images and two captions that use the same words in a different order; a metric is correct on the text task if it scores each image higher with its own caption, on the image task if it scores each caption higher with its own image, and on the group task if both hold. NegBench~\citep{alhamoud2025vision_negbench} (COCO-MCQ split) pairs each of $5{,}914$ COCO val2017 images with four candidate captions built from three templates: affirmative (\emph{``a kitchen with a table''}), negation (\emph{``a kitchen with no people''}), and hybrid, which combines both. One caption is correct and three are distractors; we score each image--caption pair independently and predict the highest-scoring caption, so chance is $25\%$.

\paragraph{Question counts.}
Pooled over benchmarks, the generator produces a mean of $2.71$ questions per unique caption (median $2$). Descriptive captions yield about five on average, from $4.16$ on Winoground to $5.52$ on RichHF (Table~\ref{tab:datasets}), and the count grows with caption length, up to $32$ for the longest RichHF caption. NegBench's templated captions each state one or two claims and yield $1.92$ on average; a few yield none, the case the $\epsilon$ in Eq.~\ref{eq:aggregation} guards against.

\subsection{Prompts}
\label{ann:prompts}

\textsc{\model} uses three prompts of its own,  one for question generation (Fig.~\ref{fig:question_generation}) and one for each answering channel (Fig.~\ref{fig:visual_answer}-~\ref{fig:text_answer}). The holistic term uses VQAScore's original prompt~\citep{vqascore}, \texttt{Does this figure show "\{caption\}"? Please answer Yes or No.}, and reads $S_{\text{hol}}$ as the probability of \textsc{yes}. All prompts run on the same backbone, and answer probabilities are read from the token distribution as described in Section~\ref{sec:impl}. The \texttt{focus}, \texttt{difficulty} and \texttt{used\_tuples} fields are logged for analysis and not used in scoring.

\begin{figure}[!htbp]
\centering
\begin{promptbox}{VISUAL ANSWER PROMPT}
You are answering a yes/no question about an image.\\

Rules:\\
1) Use only the visual evidence in the image.\\
2) Answer "yes" only if the image clearly satisfies the question; answer "no" if it clearly does not.\\

Question: \{question\}

Answer:
\end{promptbox}
\caption{\textbf{Visual answer prompt}, given the image and one question; produces $p^v_k$.}
\label{fig:visual_answer}
\end{figure}

\begin{figure}[!htbp]
\centering
\begin{promptbox}{TEXTUAL ANSWER PROMPT}
You are answering a yes/no question about what a faithful image of a text-to-image generation prompt should contain.\\

The generation prompt is the specification: it describes what a correct image must depict. Answer what that correct image should show, not just what the prompt literally spells out.\\

Rules:\\
1) Treat the prompt as the specification of a faithful image. For questions about the existence, completeness, count, or spatial relation of objects named in the prompt, answer what a correct image should show:\\
   - If a named object should be present (and, for completeness questions, fully and properly rendered), answer "yes".\\
   - If the prompt says or implies the object should NOT be present, answer "no".\\
2) Respect negations ("without", "no", "not") and explicit numbers - e.g. for "a clown without a balloon", a balloon should NOT be present, so a balloon question is "no".\\
3) Answer just yes or no.

Generation prompt:
\{prompt\}

Question: \{question\}

Answer:
\end{promptbox}
\caption{\textbf{Text-only answer prompt}, given the caption and one question without the image; produces the reference $p^t_k$.}\label{fig:text_answer}
\end{figure}

\begin{figure}[!htbp]
\centering
\begin{promptbox}{QUESTION GENERATION PROMPT}
You are checking whether an image matches a description of it. Generate yes/no questions that can each be answered by looking at the image.\\

Description:\\
\{prompt\}\\

Every question must pass the gate below. A question that fails it is noise: it will be answered unreliably and it will not tell a matching image from a mismatched one.\\

THE GATE - DECIDABLE BY LOOKING. The answer must be settled by what is visibly shown, not by what someone knows, feels, or infers. Never ask about intent, mood, atmosphere, style,
aesthetic quality, what happened before or after, or how well something is depicted. Never ask anything that needs knowledge of the world beyond the picture itself.\\

If a question fails the gate, do not keep it and do not weaken it \- replace it with a different question about the same claim that passes.\\

HOW MANY QUESTIONS: as many as the description needs. There is no target number.\\
- Every distinct thing the description names MUST have at least one question. This is a hard minimum, not a suggestion: a description naming six things gets at least six questions.\\
- Beyond that minimum, add a question for every further claim the description makes about those things - what they look like, how many there are, what they are doing, and how they are arranged relative to each other.\\
- A longer, more detailed description therefore produces more questions than a short, simple one. Never stop at a round number.\\

FLOOR: never return an empty list, and never return only one or two questions because the rest felt uncertain. Every description asserts something checkable. When you are unsure
whether a question passes the gate, KEEP IT - too few questions is a worse failure than one doubtful question.

Task:\\
- Output JSON only. Do not add markdown, prose, or any text outside the JSON.\\
- The JSON must contain a "questions" array of objects. Each object must include:\\
  - "question": a short, specific, yes/no question answerable by inspecting the image\\
  - "focus": one of ["object","count","attribute","text","spatial","action"]\\
  - "difficulty": one of ["easy","medium","hard"]\\
  - "used\_tuples": a list containing the EXACT substring(s) of the description above that this question checks. Copy the words verbatim; do not paraphrase them.\\
- Phrase every question POSITIVELY: ask whether something IS the case.\\

Return the JSON only.
\end{promptbox}
\caption{\textbf{Question-generation prompt.} \texttt{\{prompt\}} is replaced by the caption; generation runs at temperature~0.}

\label{fig:question_generation}
\end{figure}

\clearpage

\subsection{Cost}
\label{ann:cost}

\paragraph{Pass Count vs. Measured Latency.}
As shown in Table~\ref{tab:cost-passes}, \textsc{\model} requires $2N+2$ forward passes per pair ($\sim12.9$ passes on average), which is roughly double that of Soft-TIFA ($N+1$, $\sim6.5$). However, measured wall-clock latency (Table~\ref{tab:cost-latency}) shows a much smaller gap: under continuous batching, \textsc{\model} takes $0.492$\,s per pair on GenAI-Bench, only $1.36\times$ the time of Soft-TIFA ($0.362$\,s) and $1.30\times$ that of DSG ($0.379$\,s). VQAScore remains the fastest ($0.099$\,s) as it requires just one pass.

\paragraph{Why Overhead is Minimal.}
The gap between pass count and actual runtime exists because pass costs vary widely: \textbf{Question Generation Bottleneck:} Multi-token text generation for question synthesis dominates the overall runtime. This fixed overhead is shared by \textsc{\model}, Soft-TIFA, and DSG. \textbf{Single-Token Passes:} \textsc{\model}'s additional text-only answers and holistic visual pass only decode a single token each, introducing minimal computational overhead.

\paragraph{Amortization Across Multiple Images.}
When evaluating $m$ candidate images against a single prompt (e.g., best-of-$N$ selection), question generation and text-only answers are computed once and cached. In this multi-image setting, each additional image costs \textsc{\model} only $N+1$ visual passes, making its marginal cost virtually identical to Soft-TIFA ($N$ passes).

\begin{table}[h]
\centering
\small
\begin{minipage}[t]{0.55\linewidth}
\centering
\resizebox{\linewidth}{!}{%
\begin{tabular}{lcccccc}
\toprule
 & & & & \multicolumn{2}{c}{\textbf{Per pair}} & \textbf{Per extra image} \\
\cmidrule(lr){5-6}\cmidrule(lr){7-7}
\textbf{Method} & \textbf{Gen.} & \textbf{Visual} & \textbf{Text} & \textbf{Total} & \textbf{@\,$\bar N$} & \textbf{(cached)} \\
\midrule
VQAScore            & 0 & 1       & 0   & 1        & 1.0  & 1 \\
Soft-TIFA-AM        & 1 & $N$     & 0   & $N{+}1$  & 6.5  & $N$ \\
DSG                 & 3 & $N_{\mathrm{DSG}}$     & 0   & $N_{\mathrm{DSG}}{+}3$  & 9.8  & $N_{\mathrm{DSG}}$ \\
\textsc{\model}-dec & 1 & $N$     & $N$ & $2N{+}1$ & 11.9 & $N$ \\
\textbf{\textsc{\model}} & 1 & $\boldsymbol{N{+}1}$ & $\boldsymbol{N}$ & $\boldsymbol{2N{+}2}$ & \textbf{12.9} & $\boldsymbol{N{+}1}$ \\
\bottomrule
\end{tabular}}
\caption{\textbf{Forward passes.} \emph{Per pair}: one image scored against one caption, $\bar N{=}5.47$ on GenAI-Bench ($\bar N$ per benchmark in Table~\ref{tab:datasets}). \emph{Per extra image}: marginal cost when question generation and text-only answers are cached per caption. DSG writes its own questions ($\bar N_{\mathrm{DSG}}\!=\!6.85$ on GenAI-Bench).}
\label{tab:cost-passes}
\end{minipage}\hfill
\begin{minipage}[t]{0.41\linewidth}
\centering
\resizebox{\linewidth}{!}{%
\begin{tabular}{lrr}
\toprule
\textbf{Method} & \textbf{GenAI-Bench} & \textbf{5-bench.\ mix} \\
\midrule
VQAScore            & 0.099 & 0.083 \\
Soft-TIFA-AM        & 0.362 & 0.316 \\
DSG                 & 0.379 & 0.338 \\
\textsc{\model}-dec & 0.443 & 0.370 \\
\textbf{\textsc{\model}} & \textbf{0.492} & \textbf{0.411} \\
\bottomrule
\end{tabular}}
\caption{\textbf{Measured time.} Wall-clock seconds per pair on GenAI-Bench and on a mix of all five benchmarks, Qwen3.5-27B, vLLM with batching and prefix caching, one image per prompt, on a single NVIDIA B300.}
\label{tab:cost-latency}
\end{minipage}
\end{table}

\section{Additional results}
\label{ann:results}

\subsection{Full backbone sweep}
\label{ann:sweep}

\begin{figure}[h]
    \centering
    \begin{subfigure}[b]{0.8\linewidth}
        \centering
        \includegraphics[width=\linewidth]{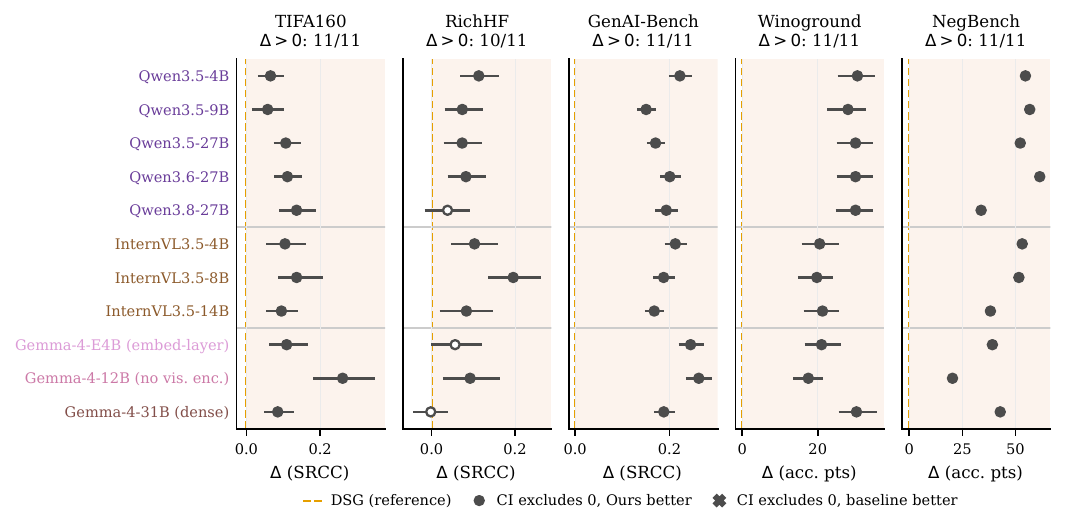}
        \caption{\textsc{\model} minus DSG}
        \label{fig:forest-dsg}
    \end{subfigure}
    \vspace{0.5em}
    \begin{subfigure}[b]{0.8\linewidth}
        \centering
        \includegraphics[width=\linewidth]{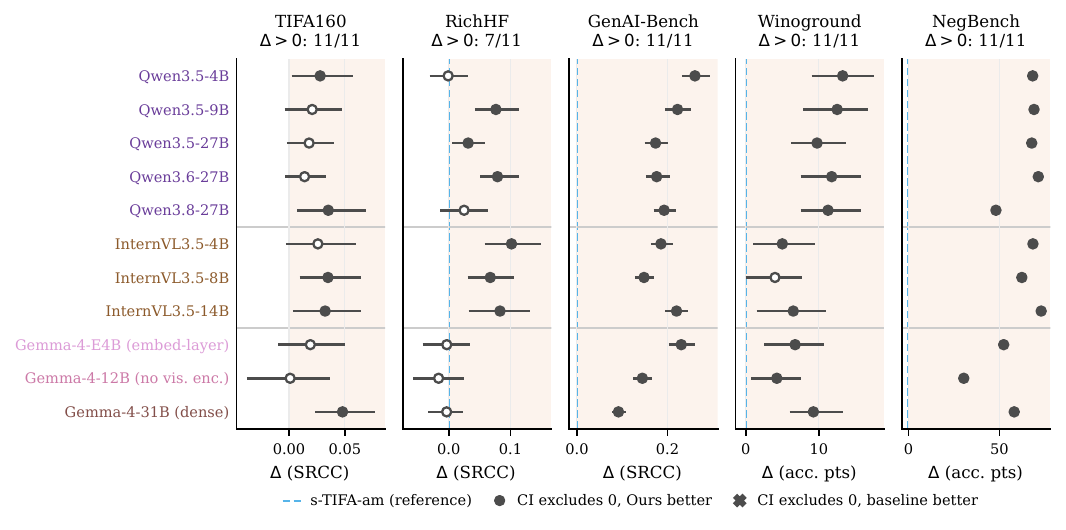}
        \caption{\textsc{\model} minus Soft-TIFA}
        \label{fig:forest-stifa}
    \end{subfigure}
    \vspace{0.5em}
    \begin{subfigure}[b]{0.8\linewidth}
        \centering
        \includegraphics[width=\linewidth]{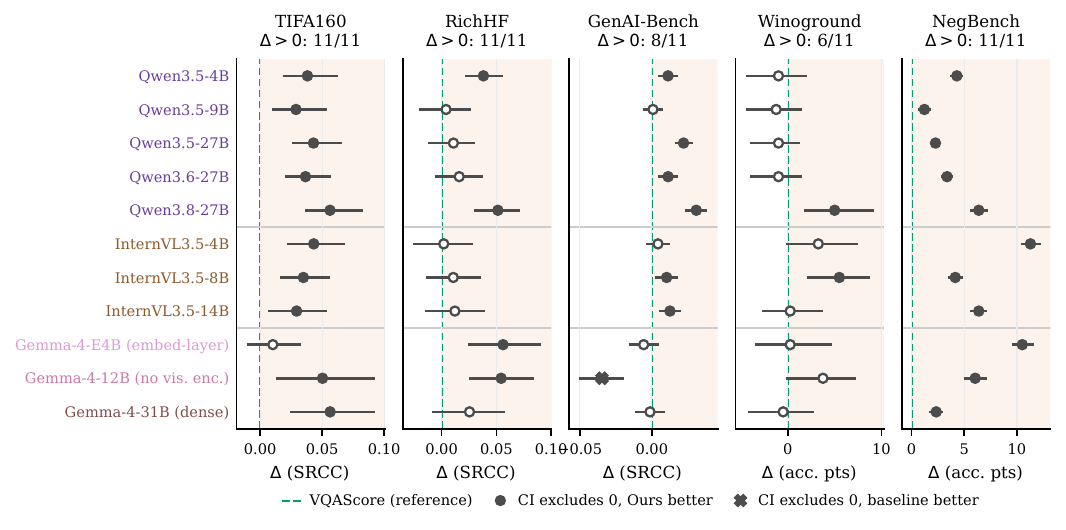}
        \caption{\textsc{\model} minus VQAScore}
        \label{fig:forest-vqa}
    \end{subfigure}
    \caption{Performance differences on every backbone and benchmark, with paired BCa 95\% intervals. Filled points exclude zero.}
    \label{fig:forest-all}
\end{figure}

Tables~\ref{tab:sweep-size} and~\ref{tab:sweep-arch-gen} report every method on all eleven backbones, grouped by the axes of Section~\ref{sec:backbones}. Fig.~\ref{fig:forest-all} repeats the paired comparison of Fig.~\ref{fig:forest} against each baseline separately.

\paragraph{Size.}
Table~\ref{tab:sweep-size} shows that both ladders behave alike. Winoground rises with size for every method, the correlation benchmarks rise modestly and not always monotonically, and \textsc{\model}'s NegBench accuracy stays flat. Its NegBench margin is largest at the smallest size in both families ($+4.3$ on Qwen3.5-4B, $+11.3$ on InternVL3.5-4B), where the holistic baseline is weakest.

\paragraph{Architecture.}
Within Gemma-4 (Table~\ref{tab:sweep-arch-gen}, top rows), the per-layer-embedding E4B tracks the weaker dense 4B model (InternVL3.5-4B) on GenAI-Bench and Winoground, and the dense 31B comes close to Qwen3.5-27B on TIFA160 and GenAI-Bench but not on RichHF. The 12B variant without a vision encoder is the weakest backbone on every benchmark, below its smaller sibling, and the only one on which \textsc{\model} loses significantly to a baseline (VQAScore on GenAI-Bench).

\paragraph{Generation.}
At a fixed 27B (Table~\ref{tab:sweep-arch-gen}, bottom rows), Qwen3.8 is a weaker answerer than Qwen3.5 and Qwen3.6 for nearly every method, yet it gives \textsc{\model} its largest margins on all five benchmarks, including the only significant Winoground gain in the sweep ($+5.0$).

\begin{table*}[!htbp]
\centering
\small
\begin{tabular}{lccccr}
\toprule
\textbf{Backbone} & \textbf{DSG} & \textbf{Soft-TIFA-AM} & \textbf{VQAScore} & \textbf{\textsc{\model}} & $\boldsymbol{\Delta}$ \\
\midrule
\multicolumn{6}{l}{\textit{TIFA160}} \\
Qwen3.5-4B      & 0.678 & \underline{0.715} & 0.705 & \textbf{0.743} & $+0.028^{*}$ \\
Qwen3.5-9B      & 0.673 & \underline{0.710} & 0.702 & \textbf{0.731} & $+0.021$ \\
Qwen3.5-27B     & 0.673 & \underline{0.762} & 0.737 & \textbf{0.780} & $+0.018$ \\
\addlinespace[2pt]
InternVL3.5-4B  & 0.592 & \underline{0.671} & 0.653 & \textbf{0.696} & $+0.026$ \\
InternVL3.5-8B  & 0.592 & \underline{0.693} & 0.693 & \textbf{0.728} & $+0.035^{*}$ \\
InternVL3.5-14B & 0.616 & 0.678 & \underline{0.681} & \textbf{0.710} & $+0.029^{*}$ \\
\midrule
\multicolumn{6}{l}{\textit{RichHF}} \\
Qwen3.5-4B      & 0.541 & \textbf{0.655} & 0.616 & \underline{0.654} & $-0.001$ \\
Qwen3.5-9B      & 0.588 & 0.585 & \underline{0.658} & \textbf{0.662} & $+0.004$ \\
Qwen3.5-27B     & 0.594 & 0.636 & \underline{0.656} & \textbf{0.667} & $+0.011$ \\
\addlinespace[2pt]
InternVL3.5-4B  & 0.457 & 0.458 & \underline{0.558} & \textbf{0.560} & $+0.002$ \\
InternVL3.5-8B  & 0.412 & 0.540 & \underline{0.597} & \textbf{0.607} & $+0.011$ \\
InternVL3.5-14B & 0.500 & 0.501 & \underline{0.572} & \textbf{0.584} & $+0.012$ \\
\midrule
\multicolumn{6}{l}{\textit{GenAI-Bench}} \\
Qwen3.5-4B      & 0.484 & 0.445 & \underline{0.695} & \textbf{0.706} & $+0.011^{*}$ \\
Qwen3.5-9B      & 0.551 & 0.479 & \underline{0.701} & \textbf{0.702} & $+0.001$ \\
Qwen3.5-27B     & 0.549 & 0.546 & \underline{0.698} & \textbf{0.720} & $+0.022^{*}$ \\
\addlinespace[2pt]
InternVL3.5-4B  & 0.415 & 0.442 & \underline{0.623} & \textbf{0.627} & $+0.004$ \\
InternVL3.5-8B  & 0.467 & 0.506 & \underline{0.644} & \textbf{0.655} & $+0.010^{*}$ \\
InternVL3.5-14B & 0.503 & 0.450 & \underline{0.658} & \textbf{0.671} & $+0.013^{*}$ \\
\midrule
\multicolumn{6}{l}{\textit{Winoground}} \\
Qwen3.5-4B      & 0.347 & 0.520 & \textbf{0.662} & \underline{0.652} & $-0.010$ \\
Qwen3.5-9B      & 0.420 & 0.575 & \textbf{0.713} & \underline{0.700} & $-0.013$ \\
Qwen3.5-27B     & 0.445 & 0.647 & \textbf{0.755} & \underline{0.745} & $-0.010$ \\
\addlinespace[2pt]
InternVL3.5-4B  & 0.138 & 0.292 & \underline{0.310} & \textbf{0.343} & $+0.033$ \\
InternVL3.5-8B  & 0.175 & \underline{0.333} & 0.318 & \textbf{0.372} & $+0.040$ \\
InternVL3.5-14B & 0.198 & 0.345 & \underline{0.407} & \textbf{0.410} & $+0.003$ \\
\midrule
\multicolumn{6}{l}{\textit{NegBench}} \\
Qwen3.5-4B      & 0.312 & 0.176 & \underline{0.816} & \textbf{0.859} & $+0.043^{*}$ \\
Qwen3.5-9B      & 0.280 & 0.156 & \underline{0.835} & \textbf{0.847} & $+0.012^{*}$ \\
Qwen3.5-27B     & 0.347 & 0.191 & \underline{0.847} & \textbf{0.869} & $+0.023^{*}$ \\
\addlinespace[2pt]
InternVL3.5-4B  & 0.265 & 0.113 & \underline{0.684} & \textbf{0.797} & $+0.113^{*}$ \\
InternVL3.5-8B  & 0.285 & 0.177 & \underline{0.759} & \textbf{0.801} & $+0.041^{*}$ \\
InternVL3.5-14B & 0.421 & 0.073 & \underline{0.739} & \textbf{0.803} & $+0.064^{*}$ \\
\bottomrule
\end{tabular}
\caption{\textbf{Size ladders}: Qwen3.5 (4B/9B/27B) and InternVL3.5 (4B/8B/14B). SRCC on the correlation benchmarks; Winoground group and NegBench accuracy as fractions. $\Delta$: \textsc{\model} minus the strongest baseline in the row; $^{*}$: paired BCa 95\% interval excludes zero (intervals in Fig.~\ref{fig:forest-all}). \textbf{Bold}: best; \underline{underline}: second.}
\label{tab:sweep-size}
\end{table*}

\begin{table*}[!htbp]
\centering
\small
\begin{tabular}{lccccr}
\toprule
\textbf{Backbone} & \textbf{DSG} & \textbf{Soft-TIFA-AM} & \textbf{VQAScore} & \textbf{\textsc{\model}} & $\boldsymbol{\Delta}$ \\
\midrule
\multicolumn{6}{l}{\textit{TIFA160}} \\
Gemma-4-E4B & 0.595 & 0.685 & \underline{0.694} & \textbf{0.704} & $+0.010$ \\
Gemma-4-12B & 0.285 & \underline{0.545} & 0.496 & \textbf{0.547} & $+0.001$ \\
Gemma-4-31B & 0.668 & \underline{0.705} & 0.697 & \textbf{0.753} & $+0.048^{*}$ \\
\addlinespace[2pt]
Qwen3.5-27B & 0.673 & \underline{0.762} & 0.737 & \textbf{0.780} & $+0.018$ \\
Qwen3.6-27B & 0.677 & \underline{0.774} & 0.752 & \textbf{0.788} & $+0.014$ \\
Qwen3.8-27B & 0.600 & \underline{0.701} & 0.680 & \textbf{0.736} & $+0.035^{*}$ \\
\midrule
\multicolumn{6}{l}{\textit{RichHF}} \\
Gemma-4-E4B & 0.519 & \textbf{0.579} & 0.519 & \underline{0.575} & $-0.003$ \\
Gemma-4-12B & 0.331 & \textbf{0.440} & 0.369 & \underline{0.423} & $-0.016$ \\
Gemma-4-31B & \underline{0.583} & \textbf{0.584} & 0.555 & 0.581 & $-0.003$ \\
\addlinespace[2pt]
Qwen3.5-27B & 0.594 & 0.636 & \underline{0.656} & \textbf{0.667} & $+0.011$ \\
Qwen3.6-27B & 0.591 & 0.594 & \underline{0.657} & \textbf{0.673} & $+0.016$ \\
Qwen3.8-27B & 0.540 & \underline{0.554} & 0.527 & \textbf{0.579} & $+0.025$ \\
\midrule
\multicolumn{6}{l}{\textit{GenAI-Bench}} \\
Gemma-4-E4B & 0.366 & 0.381 & \textbf{0.617} & \underline{0.611} & $-0.006$ \\
Gemma-4-12B & 0.165 & 0.283 & \textbf{0.462} & \underline{0.428} & $-0.035^{*}$ \\
Gemma-4-31B & 0.512 & 0.608 & \textbf{0.701} & \underline{0.700} & $-0.001$ \\
\addlinespace[2pt]
Qwen3.5-27B & 0.549 & 0.546 & \underline{0.698} & \textbf{0.720} & $+0.022^{*}$ \\
Qwen3.6-27B & 0.533 & 0.558 & \underline{0.723} & \textbf{0.734} & $+0.011^{*}$ \\
Qwen3.8-27B & 0.497 & 0.497 & \underline{0.659} & \textbf{0.690} & $+0.031^{*}$ \\
\midrule
\multicolumn{6}{l}{\textit{Winoground}} \\
Gemma-4-E4B & 0.128 & 0.270 & \underline{0.335} & \textbf{0.338} & $+0.003$ \\
Gemma-4-12B & 0.060 & 0.193 & \underline{0.198} & \textbf{0.235} & $+0.037$ \\
Gemma-4-31B & 0.420 & 0.630 & \textbf{0.728} & \underline{0.723} & $-0.005$ \\
\addlinespace[2pt]
Qwen3.5-27B & 0.445 & 0.647 & \textbf{0.755} & \underline{0.745} & $-0.010$ \\
Qwen3.6-27B & 0.458 & 0.640 & \textbf{0.767} & \underline{0.757} & $-0.010$ \\
Qwen3.8-27B & 0.383 & 0.570 & \underline{0.632} & \textbf{0.682} & $+0.050^{*}$ \\
\midrule
\multicolumn{6}{l}{\textit{NegBench}} \\
Gemma-4-E4B & 0.358 & 0.226 & \underline{0.645} & \textbf{0.750} & $+0.105^{*}$ \\
Gemma-4-12B & 0.363 & 0.262 & \underline{0.506} & \textbf{0.566} & $+0.060^{*}$ \\
Gemma-4-31B & 0.407 & 0.253 & \underline{0.812} & \textbf{0.835} & $+0.023^{*}$ \\
\addlinespace[2pt]
Qwen3.5-27B & 0.347 & 0.191 & \underline{0.847} & \textbf{0.869} & $+0.023^{*}$ \\
Qwen3.6-27B & 0.262 & 0.163 & \underline{0.843} & \textbf{0.877} & $+0.034^{*}$ \\
Qwen3.8-27B & 0.517 & 0.374 & \underline{0.792} & \textbf{0.855} & $+0.064^{*}$ \\
\bottomrule
\end{tabular}
\caption{\textbf{Architectures and generations.} Top of each block: Gemma-4 per-layer-embedding E4B, 12B without a vision encoder, and dense 31B. Bottom: Qwen3.5, 3.6 and 3.8 at a fixed 27B (Qwen3.5-27B repeated from Table~\ref{tab:sweep-size} as the reference). Columns as in Table~\ref{tab:sweep-size}.}
\label{tab:sweep-arch-gen}
\end{table*}

\clearpage

\clearpage

\subsection{Negation across backbones}
\label{ann:negbench}

Table~\ref{tab:negbench-backbones} repeats Fig.~\ref{fig:analysis}a on further backbones. The pattern holds on every one: the fixed-\textsc{yes} metrics fall below chance on negation-template items (DSG at most $0.23$, Soft-TIFA at most $0.10$) and stay near chance on hybrid ones, while \textsc{\model} keeps at least $0.78$ on negation and $0.72$ on hybrid items.

\begin{table}[!htbp]
\centering
\small
\setlength{\tabcolsep}{5pt}
\begin{tabular}{lccccc}
\toprule
\textbf{Backbone} & \textbf{DSG} & \textbf{Soft-TIFA-AM} & \textbf{VQAScore} & \textbf{\emph{Agree}} & \textbf{\textsc{\model}} \\
\midrule
\multicolumn{6}{l}{\textit{Affirmative}} \\
Qwen3.5-4B   & 0.553 & 0.485 & 0.831 & \textbf{0.888} & 0.873 \\
Qwen3.5-9B   & 0.467 & 0.427 & 0.874 & \textbf{0.911} & 0.902 \\
Qwen3.5-27B  & 0.731 & 0.483 & 0.885 & \textbf{0.912} & 0.903 \\
Qwen3-VL-4B  & 0.410 & 0.337 & 0.682 & \textbf{0.819} & 0.803 \\
Gemma-4-31B  & 0.815 & 0.432 & 0.801 & 0.796 & \textbf{0.816} \\
Gemma-4-E4B  & 0.540 & 0.454 & 0.623 & 0.716 & \textbf{0.720} \\
\midrule
\multicolumn{6}{l}{\textit{Negation}} \\
Qwen3.5-4B   & 0.191 & 0.018 & 0.830 & \textbf{0.894} & 0.872 \\
Qwen3.5-9B   & 0.091 & 0.024 & \textbf{0.835} & 0.808 & 0.827 \\
Qwen3.5-27B  & 0.107 & 0.026 & 0.844 & \textbf{0.870} & 0.862 \\
Qwen3-VL-4B  & 0.067 & 0.039 & 0.499 & \textbf{0.814} & 0.781 \\
Gemma-4-31B  & 0.072 & 0.034 & 0.880 & 0.891 & \textbf{0.900} \\
Gemma-4-E4B  & 0.230 & 0.104 & 0.721 & \textbf{0.830} & 0.824 \\
\midrule
\multicolumn{6}{l}{\textit{Hybrid}} \\
Qwen3.5-4B   & 0.180 & 0.010 & 0.787 & \textbf{0.862} & 0.833 \\
Qwen3.5-9B   & 0.266 & 0.007 & 0.795 & 0.803 & \textbf{0.812} \\
Qwen3.5-27B  & 0.181 & 0.051 & 0.810 & \textbf{0.867} & 0.843 \\
Qwen3-VL-4B  & 0.160 & 0.009 & 0.781 & 0.754 & \textbf{0.818} \\
Gemma-4-31B  & 0.306 & 0.277 & 0.759 & 0.787 & \textbf{0.794} \\
Gemma-4-E4B  & 0.294 & 0.086 & 0.595 & \textbf{0.738} & 0.722 \\
\bottomrule
\end{tabular}
\caption{\textbf{NegBench accuracy by correct-answer template} on further backbones ($n{=}5{,}914$, chance $0.25$). \emph{Agree}: the decomposed term alone; \textsc{\model}: merged with the holistic term. \textbf{Bold}: best per row. Backbones are a subset of those in Appendix~\ref{ann:sweep}.}
\label{tab:negbench-backbones}
\end{table}

\subsection{GenAI-Bench by skill}
\label{ann:skills}

Table~\ref{tab:skills} breaks GenAI-Bench down by skill tags. \textsc{\model} improves on VQAScore for every tag, significantly for all but \emph{universal}, and the gain is three times larger on prompts with an advanced skill ($+0.040$) than on basic-only prompts ($+0.013$). The largest gains are on \emph{differentiation} and \emph{negation} ($+0.067$, $+0.066$). The negation tag also shows the fixed-\textsc{yes} failure on a standard benchmark: Soft-TIFA falls to $0.062$ SRCC and DSG to $0.260$ on these prompts, against $0.601$ for \textsc{\model}.

\begin{table}[h]
\centering
\small
    \begin{tabular}{lrrrrr}
    \hline
    \textbf{Skill tag} & \textbf{DSG} & \textbf{s-TIFA-AM} & \textbf{VQAScore} & \textbf{Ours} & $\Delta$ \\ \hline
    \textit{Basic vs advanced} &  &  &  &  &  \\
    \quad Basic only (no advanced tag) & 0.591 & 0.661 & 0.685 & \textbf{0.698} & $+0.013$$^*$ \\
    \quad Advanced (any advanced tag) & 0.427 & 0.357 & 0.607 & \textbf{0.646} & $+0.040$$^*$ \\
    \textit{Basic skills} &  &  &  &  &  \\
    \quad Action Relation & 0.584 & 0.621 & 0.709 & \textbf{0.733} & $+0.024$$^*$ \\
    \quad Attribute & 0.562 & 0.594 & 0.700 & \textbf{0.726} & $+0.026$$^*$ \\
    \quad Part Relation & 0.581 & 0.560 & 0.683 & \textbf{0.715} & $+0.032$$^*$ \\
    \quad Scene & 0.529 & 0.506 & 0.706 & \textbf{0.722} & $+0.016$$^*$ \\
    \quad Spatial Relation & 0.571 & 0.610 & 0.690 & \textbf{0.720} & $+0.030$$^*$ \\
    \textit{Advanced skills} &  &  &  &  &  \\
    \quad Comparison & 0.428 & 0.521 & 0.534 & \textbf{0.579} & $+0.045$$^*$ \\
    \quad Counting & 0.472 & 0.534 & 0.596 & \textbf{0.633} & $+0.037$$^*$ \\
    \quad Differentiation & 0.423 & 0.500 & 0.498 & \textbf{0.567} & $+0.067$$^*$ \\
    \quad Negation & 0.260 & 0.062 & 0.534 & \textbf{0.601} & $+0.066$$^*$ \\
    \quad Universal & 0.501 & 0.583 & 0.685 & \textbf{0.693} & $+0.008$ \\ \hline
    \end{tabular}
    \caption{\textbf{GenAI-Bench SRCC by skill tag} (Qwen3.5-27B). A prompt can carry several tags, so item counts overlap. \textsc{\model} minus the strongest baseline; $^{*}$: paired BCa 95\% interval excludes zero.}
    \label{tab:skills}
\end{table}

\subsection{Merge-weight sensitivity}
\label{ann:lambda}

\begin{figure}[h]
    \centering
    \includegraphics[width=0.5\linewidth]{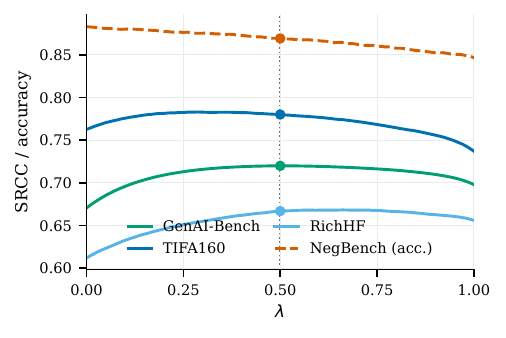}
    \caption{\textbf{Sensitivity to the merge weight} $\lambda$ (Qwen3.5-27B). $\lambda{=}0$: decomposed term only; $\lambda{=}1$: holistic term only; the dotted line marks $\lambda{=}0.5$, fixed a priori and used throughout. SRCC for the correlation benchmarks, accuracy for NegBench (dashed).}
    \label{fig:lambda}
\end{figure}

Fig.~\ref{fig:lambda} varies the weight $\lambda$ of the holistic term in Eq.~\ref{eq:merge} from $0$ (decomposed term only) to $1$ (holistic term only). On all three correlation benchmarks the merged score exceeds both terms over a broad common range, $\lambda \in [0.32, 0.82]$, and the curves are flat near their optimum: the a-priori choice $\lambda = 0.5$ is within $0.003$ SRCC of the best weight on every benchmark.

\section{Extended ablations}
\label{ann:ablations}

\subsection{Tuple extraction}
\label{ann:tuples}

Replacing direct question generation with DSG's tuple extraction is the largest loss among the pipeline components in Table~\ref{tab:ablations}. Table~\ref{tab:tuples-mechanism} shows the two mechanisms behind it.
On NegBench's negation template the extractor returns no tuples for $47\%$ of captions, the correct negated caption then scores $0$, and accuracy falls to $4.1\%$, far below chance, showing tuples cannot express absence. On Winoground captions in a pair share their entities and differ only in how they relate, so tuple-derived question sets for the two captions overlap heavily (mean Jaccard $0.44$ against $0.27$ for direct generation, $13\%$ identical), which shrinks the paired score margin and removes the contrast the benchmark tests showing tuples also cannot express binding.

\begin{table}[!htbp]
\centering
\small
\begin{minipage}[t]{0.44\linewidth}
\centering
\textbf{(a) NegBench}\\[2pt]
\begin{tabular}{lrr}
\toprule
\textbf{Template} & \textbf{Zero-tuple} & \textbf{Acc.} \\
\midrule
Affirmative & 22.7\% & 0.932 \\
Negation    & 46.8\% & 0.041 \\
Hybrid      & 23.1\% & 0.913 \\
\midrule
Overall     & 30.5\% & 0.644 \\
\bottomrule
\end{tabular}
\end{minipage}\hfill
\begin{minipage}[t]{0.54\linewidth}
\centering
\textbf{(b) Winoground}\\[2pt]
\begin{tabular}{lrr}
\toprule
 & \textbf{Direct QG} & \textbf{Tuples} \\
\midrule
Mean Jaccard (pair)    & 0.268 & 0.438 \\
Identical question sets & 0.5\% & 12.9\% \\
Median pair margin     & 0.198 & 0.118 \\
\bottomrule
\end{tabular}
\end{minipage}
\caption{\textbf{Why tuple extraction hurts} (Qwen3.5-27B). (a) Share of NegBench captions yielding no tuples, and accuracy with tuple-based questions, per template. (b) Overlap between the question sets generated for the two captions of a Winoground pair, and the median score margin between them, with direct generation vs.\ tuples.}
\label{tab:tuples-mechanism}
\end{table}

\subsection{When commitment weighting acts}
\label{ann:weighting}

The decomposed term weights each question's agreement $q_k$ by its commitment $w_k = |p^t_k - \bar p^t_k|$, which is near $1$ when the caption settles the answer and near $0$ when it leaves it open. The weighted and unweighted scores of an item differ exactly by $\operatorname{Cov}_k(w_k, a_k)/\bar w$, so weighting changes a score only when an item's questions differ in commitment, typically because one of them is left open.

Across $11$ backbones and $5$ benchmarks, weighting improves \textsc{\model} in $36$ of $55$ evaluations, worsens it in $10$ and leaves $9$ unchanged ($|\Delta| \le 0.01$), with a mean gain of $0.25$ points. Grouping each benchmark's items into quartiles by their least committed question, the gain is $0.68$ points on the quartile with the least committed questions ($34$ improved, $11$ worse) and $0.02$ on the most committed quartile, where $44$ of $55$ evaluations are unchanged. Weighting helps most on backbones whose text channel hedges more often: all three InternVL3.5 models and Qwen3.5-9B improve on every benchmark, while Qwen3.5-4B and Qwen3.5-27B lose slightly (Table~\ref{tab:cw-bb}).

\begin{table}[!htbp]
\centering
\small
\begin{tabular}{lrrr}
\toprule
\textbf{Backbone} & \textbf{Mean $\Delta$ (all items)} & \textbf{Mean $\Delta$ (least committed)} & \textbf{Improved} \\
\midrule
InternVL3.5-14B & $+1.70$ & $+4.06$ & 5/5 \\
InternVL3.5-4B  & $+0.75$ & $+1.16$ & 5/5 \\
InternVL3.5-8B  & $+0.30$ & $+1.13$ & 5/5 \\
Qwen3.5-9B      & $+0.29$ & $+0.99$ & 5/5 \\
Gemma-4-12B     & $+0.06$ & $+0.45$ & 3/5 \\
Qwen3.8-27B     & $+0.04$ & $+0.45$ & 4/5 \\
Qwen3.6-27B     & $+0.08$ & $+0.26$ & 3/5 \\
Gemma-4-31B     & $-0.02$ & $+0.13$ & 1/5 \\
Gemma-4-E4B     & $-0.01$ & $-0.14$ & 1/5 \\
Qwen3.5-4B      & $-0.13$ & $-0.22$ & 2/5 \\
Qwen3.5-27B     & $-0.27$ & $-0.79$ & 2/5 \\
\bottomrule
\end{tabular}
\caption{\textbf{Gain of \textsc{\model} from commitment weighting} per backbone, weighted minus unweighted, averaged over the five benchmarks, on all items and on the quartile of items with the least committed questions. Units: SRCC points ($\times 100$) for the correlation benchmarks, accuracy points otherwise. \emph{Improved}: benchmarks with $\Delta > 0.01$ on all items.}
\label{tab:cw-bb}
\end{table}

\subsection{Merge ablation}
\label{ann:merge-ablation}

In Table~\ref{tab:merge-ablation} we ask whether the benefit of merging depends on the decomposed term, by merging VQAScore with Soft-TIFA-AM instead of with \emph{Agree}, using the same $0.5/0.5$ recipe. The two merges tie on TIFA160 and Winoground, and the Soft-TIFA merge is slightly higher on RichHF ($-0.008$, not significant). They separate where the scoring rule matters: on GenAI-Bench the Soft-TIFA merge does not improve on VQAScore alone ($0.696$ vs.\ $0.698$), while \textsc{\model} gains $0.024$ over it, and on NegBench the Soft-TIFA merge drops to $62.2$, well below VQAScore alone ($84.7$), because the fixed-\textsc{yes} term pulls the holistic score down on exactly the negated items it gets wrong.

\begin{table}[!htbp]
\centering
\small
\begin{tabular}{llrrrrr}
\toprule
 & & & & & \textbf{\textsc{\model}} & \textbf{Soft-TIFA-AM} \\
\textbf{Benchmark} & \textbf{Metric} & \textbf{VQAScore} & \textbf{\emph{Agree}} & \textbf{Soft-TIFA-AM} & {\footnotesize (\emph{Agree}\,$\oplus$\,VQA)} & {\footnotesize $\oplus$\,VQA} \\
\midrule
GenAI-Bench & SRCC      & 0.698 & 0.671 & 0.546 & \textbf{0.720} & 0.696 \\
TIFA160     & SRCC      & 0.737 & 0.763 & 0.762 & \textbf{0.780} & \textbf{0.780} \\
RichHF      & SRCC      & 0.656 & 0.612 & 0.636 & 0.667 & \textbf{0.675} \\
Winoground  & group acc & 75.5  & 65.8  & 64.8  & 74.5  & \textbf{75.2} \\
NegBench    & MCQ acc   & 84.7  & 88.3  & 19.1  & \textbf{86.9} & 62.2 \\
\bottomrule
\end{tabular}
\caption{\textbf{Is the merge specific to \emph{Agree}?} Both merges combine a decomposed term with VQAScore using the same $0.5/0.5$ recipe; only the decomposed term differs. \textbf{Bold}: better of the two merges.}
\label{tab:merge-ablation}
\end{table}

\subsection{Reasoning in the answering step}
\label{ann:reasoning}

We evaluate enabling chain-of-thought reasoning in the answering step while holding the generated questions fixed (Table~\ref{tab:reasoning}). Reasoning makes the model commit, pushing answer probabilities toward $0$ or $1$. On GenAI-Bench and NegBench, which rank many candidates by score, this produces ties, and every reasoning variant scores at or below the reasoning-off row. Winoground, which compares only two captions per image, instead gains up to $+5.5$ group points, for two reasons. With reasoning, the holistic term less often rejects both captions of an image ($18.5\% \to 7.4\%$ of images), a case where the two low scores are nearly tied and their comparison unreliable. And with reasoning in the decomposed term, the two terms fail on more different examples (the correlation $\phi$ between their per-example errors falls from $0.52$ to $0.43$), which leaves the merge more errors to correct. We keep reasoning off to preserve graded probability estimates and avoid the extra token cost.

\begin{table}[!htbp]
\centering
\small
\begin{tabular}{llccc}
\toprule
 & \textbf{Reasoning on} & \textbf{Winoground} & \textbf{GenAI-Bench}$^{\dagger}$ & \textbf{NegBench}$^{\dagger}$ \\
 & & {\footnotesize group acc.} & {\footnotesize SRCC} & {\footnotesize accuracy} \\
\midrule
\multirow{2}{*}{Holistic term}   & none & 75.5 & \textbf{0.712} & \textbf{85.8} \\
                                 & holistic query & \textbf{78.2} & 0.664 & 84.4 \\
\midrule
\multirow{2}{*}{Decomposed term} & none & 65.7 & \textbf{0.659} & \textbf{89.7} \\
                                 & image and text answers & \textbf{66.2} & 0.607 & 85.2 \\
\midrule
\multirow{4}{*}{Merged (final)}  & none & 74.5 & \textbf{0.727} & \textbf{88.6} \\
                                 & holistic query & 77.5 & 0.705 & 86.9 \\
                                 & image and text answers & 77.5 & 0.711 & 85.8 \\
                                 & all & \textbf{80.0} & 0.677 & 86.1 \\
\bottomrule
\end{tabular}
\caption{\textbf{Chain-of-thought reasoning in the answering step} (Qwen3.5-27B), with questions held fixed so only the answering step changes. \textbf{Bold}: best per column within each block. The remaining combinations (image or text answers alone, or with the holistic query) lead to the same conclusion: none exceeds the reasoning-off row on GenAI-Bench or NegBench. $^{\dagger}$Run on a 100-prompt (600-image) GenAI-Bench subset and a 400-item NegBench subset; Winoground is the full set ($n{=}400$).}
\label{tab:reasoning}
\end{table}

\section{Analyses behind Section~\ref{sec:analysis}}
\label{ann:analysis}

\subsection{Answer distributions}
\label{ann:yesno}

In Table~\ref{tab:yesno} we report the share of unique generated questions whose caption-implied answer is \textsc{no} ($p^t < 0.5$) and the share the caption leaves undecided (commitment $w_k < 0.2$). Questions expecting \textsc{no} make up $34.7\%$ on NegBench but at most $4\%$ elsewhere. Within NegBench the share is driven by the negation template ($86.8\%$) and the hybrid template ($47.1\%$), against $0.4\%$ for affirmative captions; weighted by question count, the three reproduce the overall $34.7\%$. Undecided questions are rare ($\le 1.4\%$), consistent with the small effect of commitment weighting (Table~\ref{tab:ablations}).

\begin{table}[!htbp]
\centering
\small
\begin{tabular}{lrrr}
\toprule
\textbf{Benchmark} & \textbf{Unique Q} & \textbf{NO share} & \textbf{Undecided} \\
 & & {\footnotesize\%\,[95\% CI]} & {\footnotesize $w_k\!<\!0.2$, \%} \\
\midrule
GenAI-Bench & 8{,}753  & 3.0\,[2.6, 3.5]    & 0.3 \\
TIFA160     & 875      & 0.1\,[0.0, 0.4]    & 0.0 \\
RichHF      & 3{,}019  & 4.0\,[2.7, 5.5]    & 1.4 \\
Winoground  & 3{,}325  & 2.3\,[1.6, 2.9]    & 1.1 \\
NegBench    & 11{,}652 & 34.7\,[33.8, 35.6] & 0.2 \\
\bottomrule
\end{tabular}
\caption{\textbf{Answer distributions per benchmark} (Qwen3.5-27B). Shares are over unique (caption, question) pairs: \emph{NO share}, $p^t < 0.5$; \emph{undecided}, commitment $w_k < 0.2$. Intervals from a caption-level bootstrap; NegBench is restricted to correct captions, Winoground to matched pairings.}
\label{tab:yesno}
\end{table}

\subsection{Reference reliability}
\label{ann:reference}

We check the polarity of the caption-only reference $p^t$ on NegBench captions containing a negation, where the absent and present entities are known. Parsing these captions (covering $80.1\%$ of them) and attributing each question to the negated or an affirmed entity, the text channel answers \textsc{yes} to all $3{,}042$ affirmed-entity questions and \textsc{no} to $92.1\%$ of the $9{,}454$ negated-entity ones. Of the rest, $443$ are phrased negatively (\emph{``Is there no person in the image?''}), where \textsc{yes} is correct, giving $96.8\%$ for negated entities and $97.6\%$ overall. The remaining $304$ ($3.2\%$) count as errors, though some may be parser misattributions.

\subsection{Score distributions}
\label{ann:calib}

\begin{figure}[h]
    \centering
    \includegraphics[width=0.6\linewidth]{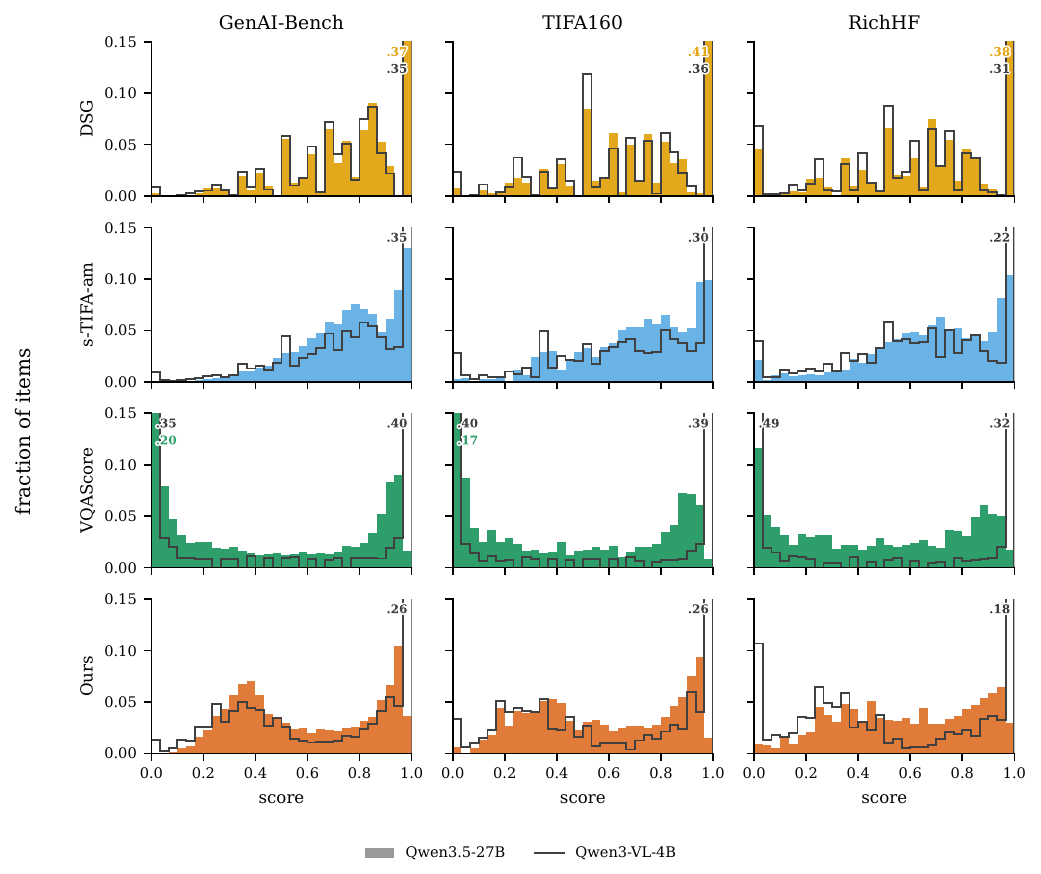}
   \caption{\textbf{Score distributions on the three human-correlation benchmarks.} Fraction of items per score bin for each metric on Qwen3.5-27B (filled) and Qwen3-VL-4B (outline); bars exceeding the axis are labelled with their height.}
    \label{fig:calib}
\end{figure}

Fig.~\ref{fig:calib} shows how the four metrics distribute their scores. VQAScore is near-binary on Qwen3-VL-4B, with most items in its two end bins, and the fixed-\textsc{yes} metrics pile up at the top: DSG places $31$--$41\%$ of items at exactly $1.0$ on both backbones, leaving them unrankable. \textsc{\model} is the most evenly spread in every panel, with $60$--$64\%$ of its mass between $0.2$ and $0.8$ on Qwen3.5-27B. Its gain over VQAScore comes from the low end: its lowest-scored $5\%$ of items are rated worse by humans than VQAScore's on every benchmark and backbone, while the top tails coincide.

\section{Qualitative examples}
\label{ann:qualitative}

Figs.~\ref{fig:example-1}--\ref{fig:negbench-2} show qualitative results for \textsc{\model} on Qwen3.5-27B. Figs.~\ref{fig:example-1}--\ref{fig:example-2} trace the full pipeline on individual image--caption pairs. The remaining figures cover the two selection tasks, choosing the image that best matches a caption (Figs.~\ref{fig:winoground}--\ref{fig:genaibench-2}) and the caption that best matches an image (Figs.~\ref{fig:negbench-1}--\ref{fig:negbench-2}), alongside VQAScore, Soft-TIFA, DSG, and the human judgments.

In these examples, \textsc{\model} selects the human-preferred image or the correct caption where the baselines do not, most clearly on explicit negation (Fig.~\ref{fig:negbench-2}, where only \textsc{\model} selects the correct caption) and on attribute and spatial binding (Figs.~\ref{fig:genaibench-1}--\ref{fig:genaibench-2}). Its scores are also graded, whereas DSG often assigns a tied score of $1.0$ to several candidate images.

Fig.~\ref{fig:fail-examples} shows two failures, one per selection task, and the per-question scores expose the cause of each. On NegBench~\#2034 the negation is handled correctly, since the text channel expects \emph{no bottle} for the correct caption, but the image channel wrongly sees a bottle ($p^v \approx 0.7$). This supports the distractor \emph{``This image features a bottle''} and contradicts the correct \emph{``A bottle is not shown''}, and because the holistic term makes the same perceptual error ($0.41$ against $0.15$), \textsc{\model} ranks the correct caption third of four ($0.22$ against $0.53$). On GenAI-Bench~\#247, all methods prefer the Midjourney image, rated $4.0$ by humans, over the DALL-E~3 image rated $5.0$. Both terms of \textsc{\model} favour it: the holistic term scores it $0.95$ against $0.69$, and in the decomposed term the two images differ mainly on whether the scroll is lit by the crystals, which the image channel affirms for Midjourney (agreement $0.98$) but largely rejects for DALL-E~3 ($0.35$), while neither image is judged to show exactly two crystals. The two images are close in faithfulness, and the one-point gap in human ratings may reflect rating noise as much as a metric error.

\begin{figure}[h]
    \centering
    \resizebox{\textwidth}{!}{\input{figures_tikz/figures/fig_qual_examples_main_1.tikz}}
    \caption{Qualitative Results with scores per questions.}
    \label{fig:example-1}
\end{figure}

\begin{figure}[h]
    \centering
    \resizebox{\textwidth}{!}{\input{figures_tikz/figures/fig_qual_examples_main_2.tikz}}
    \caption{Qualitative Results with scores per questions.}
    \label{fig:example-2}
\end{figure}

\begin{figure}[h]
    \centering
    \resizebox{\linewidth}{!}{\input{figures_tikz/figures/fig_qual_scores_main.tex}}
    \caption{More examples.}
    \label{fig:winoground}
\end{figure}

\begin{figure}[h]
    \centering
    \resizebox{\textwidth}{!}{\input{figures_tikz/figures/fig_qual_ranking_main_1.tikz}}
    \caption{GenAI-Bench examples with 6 image, 4 scorers and human evaluation.}
    \label{fig:genaibench-1}
\end{figure}

\begin{figure}[h]
    \centering
    \resizebox{\textwidth}{!}{\input{figures_tikz/figures/fig_qual_ranking_main_2.tikz}}
    \caption{GenAI-Bench examples with 6 image, 4 scorers and human evaluation.}
    \label{fig:genaibench-2}
\end{figure}

\begin{figure}[h]
    \centering
    \resizebox{\textwidth}{!}{\input{figures_tikz/figures/fig_qual_negbench_main_1.tikz}}
    \caption{Negbench example with four candidate captions.}
    \label{fig:negbench-1}
\end{figure}

\begin{figure}[h]
    \centering
    \resizebox{\textwidth}{!}{\input{figures_tikz/figures/fig_qual_negbench_main_2.tikz}}
    \caption{Negbench example with four candidate captions.}
    \label{fig:negbench-2}
\end{figure}

\begin{figure}[h]
    \centering
    \resizebox{\textwidth}{!}{\input{figures_tikz/figures/fig_qual_examples_fail_main.tikz}}
    \caption{Question scoring on the failed examples.}
    \label{fig:fail-examples}
\end{figure}

\end{document}